\documentclass[dvipsnames]{article} %
\usepackage{main}

\usepackage{booktabs}
\usepackage{enumitem}
\usepackage{wrapfig}
\usepackage{algorithm}
\usepackage{algpseudocode}
\usepackage{graphicx}
\usepackage[misc]{ifsym}
\usepackage{multicol} 
\usepackage{microtype}
\usepackage{amsmath}
\usepackage{colortbl}
\usepackage[utf8]{inputenc}
\usepackage[T1]{fontenc}
\definecolor{lightgray}{rgb}{0.9,0.9,0.9}
\usepackage{caption}
\usepackage{subcaption}
\usepackage{graphicx}
\usepackage{subcaption}
\usepackage{setspace}
\usepackage{url}
\usepackage{multirow}
\usepackage{tabularx}
\usepackage{blindtext}
\usepackage{pgfplots}
\usepackage{tikz}
\usetikzlibrary{er,positioning,bayesnet}
\usepackage{makecell}
\usepackage{tipa}
\usepackage{siunitx}
\usepackage{nicefrac}
\usepackage{listings}
\usepackage[raster,skins, most]{tcolorbox} %
\usepackage{titletoc}
\usepackage{xltabular}
\usepackage{adjustbox}
\usepackage{xurl}
\usepackage{rotating}
\usepackage[normalem]{ulem}

\useunder{\uline}{\ul}{}

\usepackage{amsmath,amsfonts,bm}

\def\eqref#1{equation~\ref{#1}}

\def\1{\bm{1}}

\DeclareMathAlphabet{\mathsfit}{\encodingdefault}{\sfdefault}{m}{sl}
\SetMathAlphabet{\mathsfit}{bold}{\encodingdefault}{\sfdefault}{bx}{n}

\renewcommand{\ghlink}{https://github.com/AQ-MedAI/SimpleSearch-VL}
\usepackage{etoolbox}
\newcommand{\modelscopelogo}{\raisebox{-1.5pt}{\includegraphics[height=1.05em]{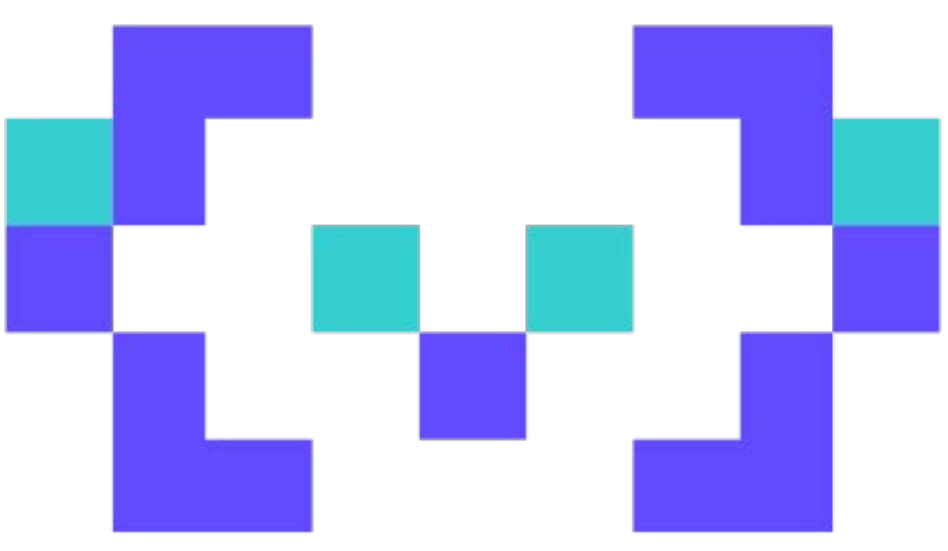}}}
\newcommand{\projectlinks}{%
  \href{\ghlink}{\github\enspace GitHub}\qquad
  \href{https://www.modelscope.cn/collections/dmmm997/SimpleSearch-VL-Data}{\modelscopelogo\enspace Dataset}\qquad
  \href{https://www.modelscope.cn/collections/dmmm997/SimpleSearch-VL}{\modelscopelogo\enspace Models}%
}
\makeatletter
\patchcmd{\@maketitle}{\github & \url{\ghlink}}{\multicolumn{2}{c}{\projectlinks}}{}{\PackageError{main}{Could not update title resource links}{Check the title link block in main.sty.}}
\makeatother

\newcommand*\justify{%
  \fontdimen2\font=0.4em
  \fontdimen3\font=0.2em
  \fontdimen4\font=0.1em
  \fontdimen7\font=0.1em
  \hyphenchar\font=`\-
}

\renewcommand{\texttt}[1]{%
  \begingroup
  \ttfamily
  \begingroup\lccode`~=`/\lowercase{\endgroup\def~}{/\discretionary{}{}{}}%
  \begingroup\lccode`~=`[\lowercase{\endgroup\def~}{[\discretionary{}{}{}}%
  \begingroup\lccode`~=`.\lowercase{\endgroup\def~}{.\discretionary{}{}{}}%
  \catcode`/=\active\catcode`[=\active\catcode`.=\active
  \justify\scantokens{#1\noexpand}%
  \endgroup
}

\newcommand{\gain}[1]{\raisebox{-0.35ex}{\scriptsize\textcolor{red!70!black}{\,+#1}}}
\newcommand{\drop}[1]{\raisebox{-0.35ex}{\scriptsize\textcolor{green!55!black}{\,-#1}}}
\newcommand{\nochg}{\raisebox{-0.35ex}{\scriptsize\textcolor{gray}{\,0.0}}}
\usepackage{makecell}
\usetikzlibrary{tikzmark}
\makeatletter
\newcommand*\myfontsize{%
  \@setfontsize\myfontsize{7}{8}%
}
\makeatother

\definecolor{uclablue}{RGB}{159, 195, 224}

\definecolor{uclagold}{RGB}{255, 240, 180}

\definecolor{aliceblue}{RGB}{255, 238, 241}

\definecolor{cadmiumgreen}{rgb}{0.0, 0.42, 0.24}

\definecolor{myred}{rgb}{0.7, 0.3, 0.0}
\definecolor{myblue}{rgb}{0.2, 0.3, 0.6}
\definecolor{babygreen}{rgb}{0.85, 0.97, 0.85}

\definecolor{purple1}{RGB}{126, 107, 196}
\definecolor{purple2}{RGB}{199, 158, 207}
\definecolor{purple3}{RGB}{214, 200, 255}
\definecolor{purple4}{RGB}{254, 240, 255}

\definecolor{deepblue}{RGB}{48, 58, 82}

\titlecontents{section}
  [1.6em]
  {\addvspace{6pt}\bfseries}
  {\contentslabel[\thecontentslabel]{1.6em}}
  {\hspace*{-1.6em}}
  {\hfill\contentspage}
\titlecontents{subsection}
  [4em]
  {\addvspace{2pt}\normalfont\small}
  {\contentslabel[\thecontentslabel]{2.4em}}
  {\hspace*{-2.4em}}
  {\titlerule*[6pt]{.}\contentspage}

\definecolor{deepPurple}{HTML}{330066}

\definecolor{uclablue_old}{rgb}{0.15, 0.45, 0.68}
\hypersetup{
    breaklinks,
    citecolor=uclablue_old,
    colorlinks=true,
    linkcolor=red
}

\newtcolorbox{mybox}[2][]
  {colback = black!5!white, colframe = black!75!black, fonttitle = \bfseries,
    colbacktitle = black!100!black, enhanced, before upper={\fontsize{8}{11}\obeyspaces\obeylines\selectfont}, fontupper=\selectfont,
    attach boxed title to top left={yshift=-2.2mm,xshift=4mm},
    title=#2,#1}

\title{%
\begin{tabular}[t]{l} 
  \parbox[t]{0.8\textwidth}{\centering 
    SimpleSearch-VL: A Simple Recipe for Multimodal Agentic Deep Search
  }
\end{tabular}
}

\author{
\large Ming Dai\textsuperscript{1,2}, Zhihong Lu\textsuperscript{2}, Jinjie Gu\textsuperscript{2}, Jiedong Zhuang\textsuperscript{3}, Yefeng Liu\textsuperscript{2}, Jing Hao\textsuperscript{2}, Lijun Wu\textsuperscript{2}, Wankou Yang\textsuperscript{1,*}, Jian Wang\textsuperscript{2,*}, Chunhua Shen\textsuperscript{2,3}.
  \\[1em]               
  {\fontsize{10pt}{11pt}\selectfont          
\textsuperscript{1}Southeast University \quad \textsuperscript{2}Ant Group \quad \textsuperscript{3}Zhejiang University}\\
}

\begin{document}

\maketitle
\begingroup
\renewcommand{\thefootnote}{\fnsymbol{footnote}}
\footnotetext[1]{Corresponding authors: Jian Wang (bobblair.wj@antgroup.com) and Wankou Yang (wkyang@seu.edu.cn).}
\endgroup

\begin{abstract}
We present \textbf{SimpleSearch-VL}, an efficient, reliable, and practical framework for multimodal agentic search. Its core idea is to improve the agent's own search-and-verification process rather than scaling data, tools, or auxiliary model components. For efficiency, \emph{Factorized Adaptive Rollout} (FAR) improves sampling efficiency by forming more informative training groups while using redundant samples to mitigate long-tail latency and expose hard samples. For reliability, SimpleSearch-VL performs \emph{evidence-verified reasoning}, explicitly using chain-of-thought verification to assess the relevance of retrieved visual and textual cues to the original context. For practicality, SimpleSearch-VL keeps a lightweight tool interface and performs webpage self-summary within the agent, requiring no additional external dependencies. With only 5K supervised tool-interleaved trajectories and 2K RL data, SimpleSearch-VL improves Qwen3-VL agentic baselines by 15.8 and 16.0 average points for the 8B and 30B-A3B variants, respectively. The SimpleSearch-VL-30B-A3B model further achieves performance competitive with agentic Gemini-3-Pro.
\end{abstract}

\begin{figure}[H]
    \centering
    \includegraphics[width=1.0\linewidth]{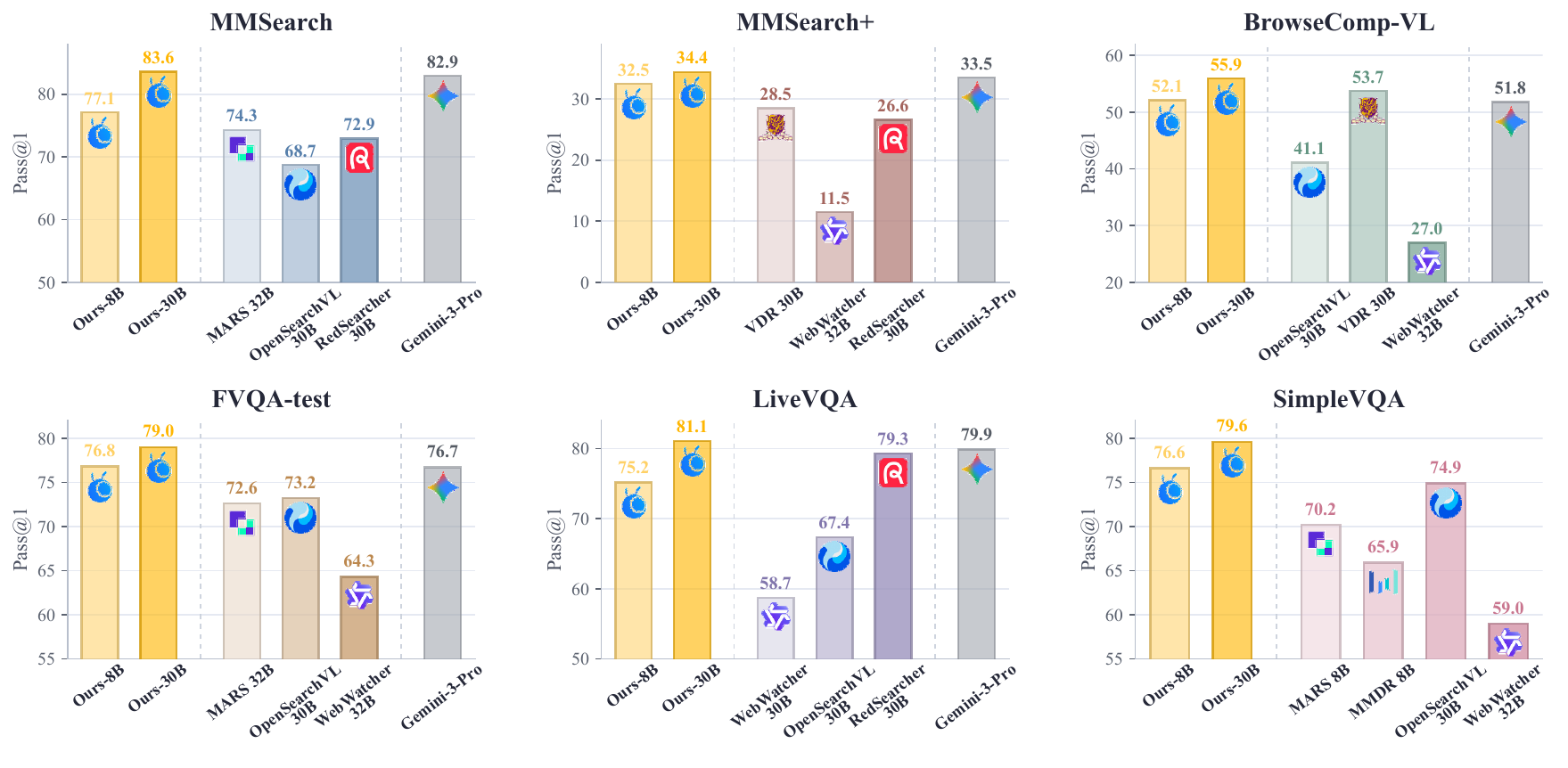}
    \vspace{-25pt}
    \caption{\textbf{Performance comparison with representative multimodal deep search agents.} SimpleSearch-VL-8B outperforms most open-source 30B-scale multimodal deep search agents, while SimpleSearch-VL-30B-A3B achieves performance competitive with agentic Gemini-3-Pro.}
    \label{fig:main-performance-comparison}
\end{figure}

\vfill

\section{Introduction}
\label{sec:intro}

\begin{figure}[t]
    \centering
    \includegraphics[width=1.0\linewidth]{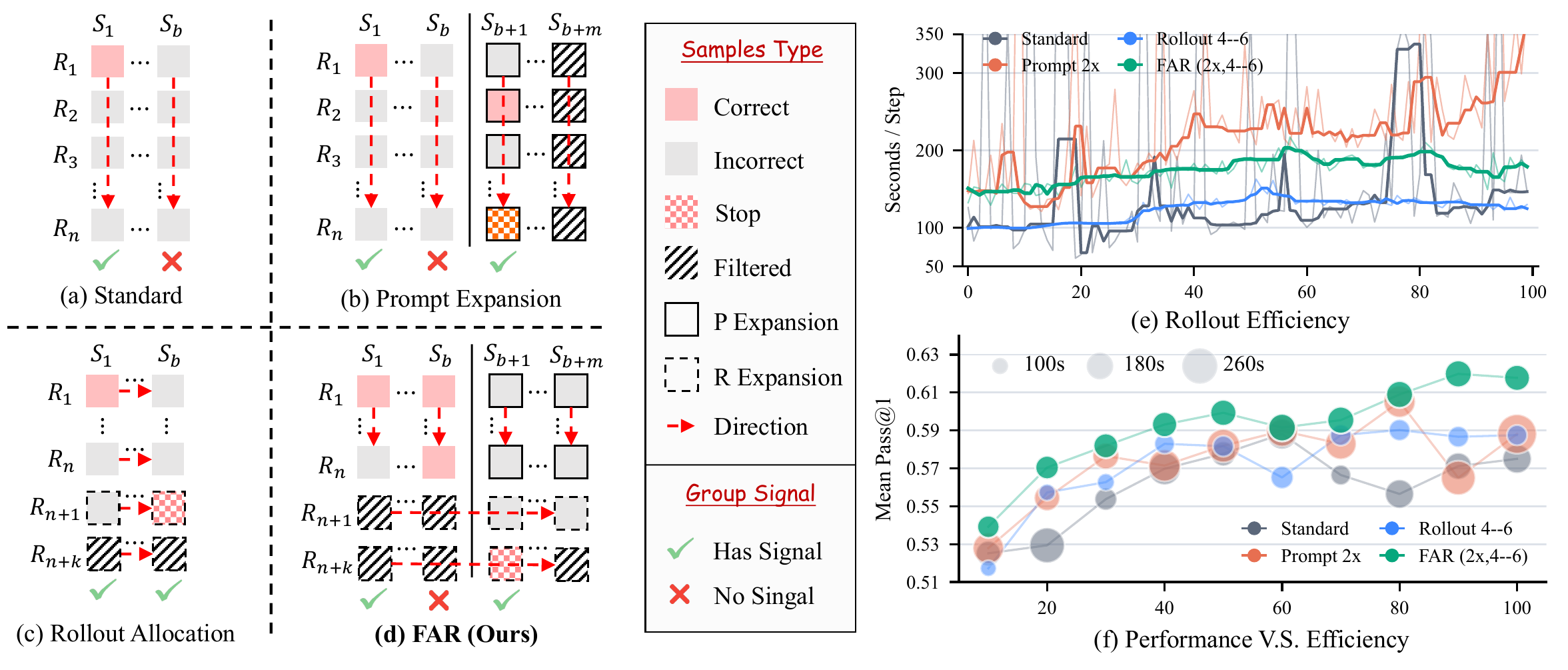}
    \caption{
    \textbf{Overview of Factorized Adaptive Rollout.}
    (a)--(d) compare fixed sampling, prompt expansion, rollout allocation, and FAR, which keeps exploring hard groups while skipping redundant tail rollouts.
    (e)--(f) show rollout time and the accuracy--efficiency trade-off.
    }
    \label{fig:far}
    \vspace{10pt}
\end{figure}

Agentic search extends large language models from passive response generation to active evidence acquisition.
In the language-only setting, recent Deep Search agents have shown that models can answer knowledge-intensive questions by issuing web-search queries, visiting webpages, coding, and synthesizing evidence across multiple sources~\citep{team2025tongyi,mirothinker,du2026openseeker,li2026openresearcher}.
Recent multimodal search agents extend this paradigm to knowledge-intensive visual reasoning by equipping MLLMs with external visual lookup and visual tool use~\citep{mmsearch-r1,webwatcher,peng2026mta,deepeyesv2,chng2025sensenova,li2026hypereyes}.
This setting is increasingly important for benchmarks that require external, up-to-date, or entity-specific visual knowledge~\citep{chen2023can,mmsearch,fu2025livevqa,cheng2025simplevqa}.

However, multimodal search agents remain difficult to make \textbf{efficient}, \textbf{reliable}, and \textbf{practical}. \textbf{Efficiency} is constrained by long-tail rollout generation: each trajectory may involve multi-turn external search and webpage visits, so a small number of samples can consume most rollout-generation time. Fixed rollout budgets further cannot adaptively terminate sampling for prompts that already provide useful reward variation or reallocate attempts to hard prompts that still lack such signal. \textbf{Reliability} requires evidence that is grounded and checkable across sources and modalities, since retrieved text, webpages, or image-search titles and URLs can appear plausible yet remain unsupported or mismatched. \textbf{Practicality} favors reproducible systems that avoid complex tool orchestration and unnecessary model dependencies. SimpleSearch-VL addresses these challenges with factorized adaptive rollout, evidence-verified reasoning, and self-summarized visit.

\textbf{Rollout efficiency.} Reasoning-oriented RL relies on multiple responses per prompt to estimate relative advantages~\citep{shao2024deepseekmath}. In multimodal agentic search, this rollout bottleneck is amplified because trajectories are slowed not only by model reasoning, but also by waiting for external tools. As shown in Fig.~\ref{fig:far}(a), fixed-size group sampling~\citep{guo2025deepseek} must wait for all assigned rollouts to finish, so a few slow tool-interleaved trajectories can dominate the wall-clock time of a training step. Prior work improves rollout efficiency along different axes: DAPO-style dynamic sampling~\citep{yu2026dapo} changes the prompt-group pool, as illustrated in Fig.~\ref{fig:far}(b), while partial or asynchronous rollout scheduling reduces long-tail stalls by interrupting, buffering, resuming, or repacking trajectories~\citep{zhou2025april,qu2025copris,sheng2026laminar}; adaptive response allocation changes the number of responses per prompt, as illustrated in Fig.~\ref{fig:far}(c)~\citep{zhang2025improving,zhang2026train,nguyen2026adaptive}. In contrast, \textbf{Factorized Adaptive Rollout} (FAR) combines the two budget dimensions in Fig.~\ref{fig:far}(d): Prompt Expansion introduces new prompt groups when more useful signal is needed, while Rollout Allocation gives additional attempts to groups that still lack reward contrast and skips redundant tail rollouts once a signal is obtained. FAR thus turns fixed-budget rollout generation into signal-aware allocation, without adding system complexity.

\textbf{Evidence verification.} Search tools provide agents with external candidate information, but strong agentic search also requires the model to select evidence that is relevant, supported, and usable for the current question. Text-based search agents can often treat retrieved passages as directly inspectable evidence~\citep{team2025tongyi,jin2025searchr1trainingllmsreason}. In multimodal search, evidence verification is more delicate because a reverse-image-search candidate may be internally coherent while still failing to match the queried image or region. As illustrated in the middle panel of Fig.~\ref{fig:markov}, a returned candidate can provide a plausible thumbnail, title, and source URL for a visually similar but different object. Although recent multimodal agents improve tool use through supervised trajectories or RL~\citep{mmsearch-r1,webwatcher,deepeyesv2}, this visual-to-source validation step is often left implicit. SimpleSearch-VL exposes it directly: the \texttt{image\_search} tool returns the matched thumbnail, webpage title, and source URL, enabling the model to verify whether the candidate matches the queried visual content before using its title, URL, or webpage evidence. Since the thumbnail is small and used only for consistency checking, it adds little token overhead while making retrieved visual evidence directly checkable.

\textbf{Tool-interface simplicity.}
Recent multimodal search agents often implement tool use as coupled pipelines, such as binding search to webpage visiting or chaining reverse image search to subsequent browsing~\citep{chng2025sensenova,vdr}. While effective, these designs can collect broad evidence bundles before the model has decided what information is needed, increasing inference cost and reducing the agent's control over subsequent evidence seeking. Some systems also rely on an external webpage summarizer, which introduces an additional model dependency during training and evaluation~\citep{mmsearch-r1,chng2025sensenova}. SimpleSearch-VL instead exposes decoupled evidence actions: at each step, the model decides which links to visit and summarize, and for multi-image inputs, which image and region should serve as the reverse-image-search query. As shown in the right panel of Fig.~\ref{fig:markov}, each webpage visit is followed by goal-conditioned self-summary, where the agent extracts query-relevant information without calling a separate summarization model. This keeps webpage understanding inside the agent while reducing tool orchestration and model dependencies.

Together, these designs make SimpleSearch-VL an \textbf{efficient}, \textbf{verifiable}, and \textbf{practical} framework for multimodal agentic search. It is also data-efficient and low-cost to train: instantiated with Qwen3-VL~\citep{Qwen3-VL}, SimpleSearch-VL learns strong search behavior from 5K supervised tool-interleaved trajectories and 2K RL data, with each variant trained on 8 H200 GPUs in about one day. Despite this modest training scale, the 8B and 30B-A3B variants improve their corresponding Qwen3-VL agentic baselines by 15.8 and 16.0 average points, and the 30B-A3B model reaches performance competitive with agentic Gemini-3-Pro on shared benchmarks.

Our main contributions are summarized as follows:
\begin{itemize}[leftmargin=*]
    \item We present SimpleSearch-VL, an efficient, verifiable, and practical framework for training multimodal search agents with a lightweight tool interface and reduced dependence on external models.
    \item We propose Factorized Adaptive Rollout, a signal-aware allocation strategy that expands prompt coverage while using redundant samples to mitigate long-tail stalls and expose hard samples.
    \item We introduce visual evidence verification with thumbnail-aware reverse image search, enabling the agent to check whether retrieved evidence matches the queried visual content before using it.
    \item We validate SimpleSearch-VL on six multimodal search benchmarks, where the 8B and 30B-A3B variants improve agentic baselines by 15.8 and 16.0 average points.
\end{itemize}

\section{Related Work}
\label{sec:related-work}

\subsection{Rollout Sampling for Reinforcement Learning}

Rollout generation is a central efficiency bottleneck in RL post-training, from PPO-style on-policy optimization~\citep{schulman2017proximal} to reasoning-oriented RL~\citep{shao2024deepseekmath} and further to agentic RL with multi-turn tool calls~\citep{dong2025arpo}. Recent studies treat rollout as an explicit sampling problem, covering generation, filtering, control, and replay mechanisms~\citep{surana2026generate}. Dynamic-rollout methods improve useful-signal density by filtering or selecting informative prompt groups, from dynamic sampling~\citep{yu2026dapo} to pre-rollout selective sampling and variance-aware sample selection~\citep{zheng2026act,hu2025vade}; complementary work instead extracts signal from zero-variance prompts rather than discarding them~\citep{le2025no}. Partial-rollout systems target wall-clock inefficiency from long-tail generations by interrupting, buffering, and resuming unfinished trajectories~\citep{zhou2025april,qu2025copris}, while asynchronous rollout systems further reduce synchronization stalls through trajectory-level scheduling and repacking~\citep{sheng2026laminar}. Adaptive-rollout methods allocate different numbers of responses across prompts according to difficulty, uncertainty, or expected gradient variance~\citep{zhang2025improving,zhang2026train,nguyen2026adaptive}. In contrast, as illustrated in Fig.~\ref{fig:far}, Factorized Adaptive Rollout (FAR) shifts budget toward hard groups that still lack useful reward variation, while skipping redundant tail rollouts once a training signal has been obtained. Because FAR generates candidate redundancy at the group level, completed alternative rollouts can provide the needed signal instead of waiting for slow or stalled tool-interleaved trajectories.

\subsection{Multimodal Deep Search Agents}

Multimodal deep search agents tackle knowledge-intensive visual questions by actively gathering and verifying external evidence through multi-turn tool interactions. Compared with standard RAG~\citep{lewis2020retrieval,yu2024visrag}, they move beyond fixed retrieval by adaptively deciding what to inspect, which modality to query, and when to refine the search. Recently, some methods construct VQA tasks paired with supervised multi-hop web-search trajectories to teach search behavior~\citep{webwatcher,deepmmsearch-r1,chen2026opensearch,peng2026mta}. Another line scales long-horizon search by training agents to decompose complex questions and iteratively gather evidence over extended tool-use trajectories~\citep{huang2026vision,chu2026redsearcher,du2026towards}. Related tool-policy work studies how models should invoke and coordinate external tools, including visual operations, retrieval tools, native agentic search behavior, and efficiency-aware parallel search~\citep{deepeyesv2,chng2025sensenova,li2026hypereyes}. Other efforts shift the target output or system bottleneck, such as grounded multimodal long-report generation and visual perception~\citep{ye2026deep,yang2026web}, or address the growing interaction history through context compression~\citep{liu2026points}. SVRL~\citep{sathish2026svrl} explicitly learns to verify and filter retrieved evidence through reinforcement learning, while ProMMSearchAgent~\citep{yan2026prommsearchagent} studies process-oriented rewards and transfer from a static training sandbox to live search. On-policy Data Evolution (ODE)~\citep{huang2026onpolicy} refines a data generator using policy-rollout feedback. These approaches motivate a more targeted examination of visual evidence verification, particularly whether retrieved visual candidates match the queried image or region. To address this limitation, we propose evidence-verified reasoning to improve textual and visual evidence verification, together with a self-summary mechanism that enables the agent to summarize webpages autonomously without relying on external models.

\section{SimpleSearch-VL}

\begin{figure}[t]
    \centering
    \includegraphics[width=1.0\linewidth]{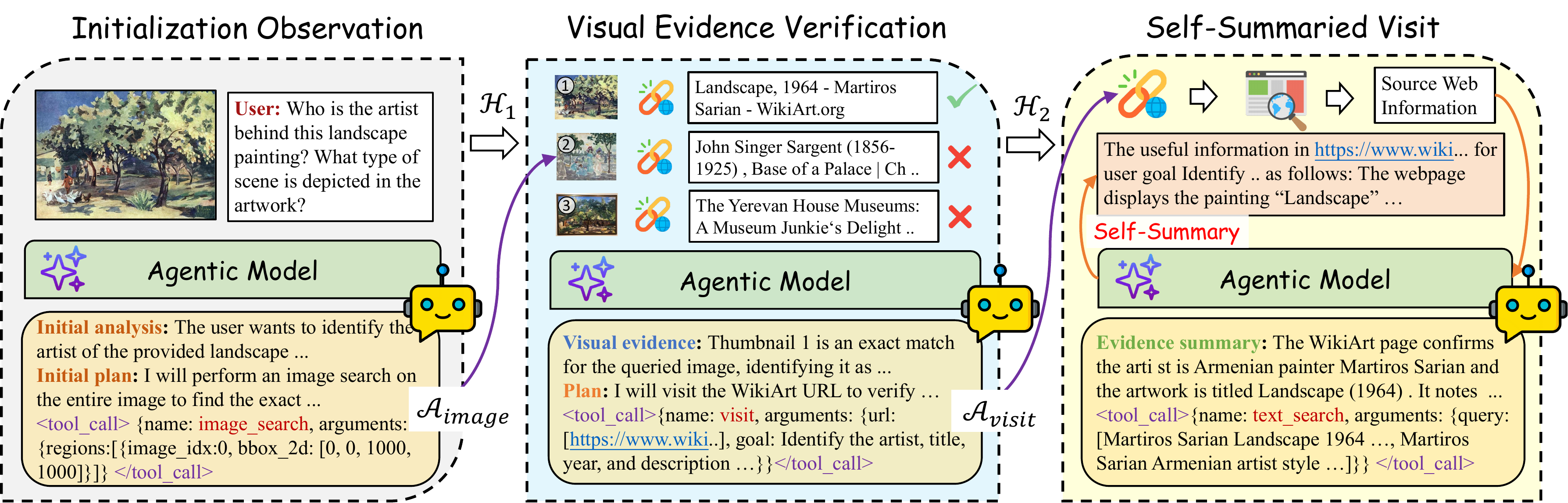}
    \caption{\textbf{Agentic search process of SimpleSearch-VL.} The model alternates between reasoning, tool calls, evidence verification, and final answer generation.}
    \label{fig:markov}
    \vspace{10pt}
\end{figure}

\subsection{Preliminaries and Overview}
\label{subsec:overview}

Given a visual question \(Q\) and input images
\(\mathcal{I}=\{I_m\}_{m=0}^{M-1}\), SimpleSearch-VL casts multimodal search as
a Markov decision process. As illustrated in
Fig.~\ref{fig:markov}, the agent alternates between reasoning,
tool use, evidence verification, and answer generation.
\((\mathcal{S},\mathcal{A},\mathcal{T},\mathcal{R})\):

\begin{itemize}[leftmargin=*, nosep, itemsep=2pt]
    \item \textbf{State $s_t \in \mathcal{S}$:} The state contains the question,
    input images, and accumulated interaction history \(\mathcal{H}_t\):
    \begin{equation}
        s_t=(Q,\mathcal{I},\mathcal{H}_t), \qquad
        \mathcal{H}_t=\{(a_1,o_1),\ldots,(a_{t-1},o_{t-1})\}.
        \label{eq:agent-state}
    \end{equation}
    Here \(a_i\) is a previous action and \(o_i\) is the corresponding tool
    observation. Conditioning on \(\mathcal{H}_t\) makes the state contain all
    information needed for the next decision.

    \item \textbf{Action $a_t \in \mathcal{A}$:} The policy first generates
    \texttt{<thinking>} and then emits exactly one action:
    \begin{equation}
        a_t \sim \pi_\theta(\cdot \mid s_t), \qquad
        \mathcal{A}=\{\mathcal{A}_{\mathrm{text}},
        \mathcal{A}_{\mathrm{image}},\mathcal{A}_{\mathrm{visit}},
        \mathcal{A}_{\mathrm{answer}}\}.
    \end{equation}
    These actions correspond to text search, reverse image search over selected
    regions, webpage visit, and final answer generation.

    \item \textbf{Transition $\mathcal{T}$:} If \(a_t\) is a tool call,
    \(\mathcal{T}\) returns an observation \(o_t=\mathcal{T}(s_t,a_t)\) and
    appends \((a_t,o_t)\) to the history. Text search returns snippets and candidate URLs;
    \texttt{image\_search} returns matched thumbnails, webpage titles, and
    source URLs from reverse image search over queried regions;
    visit returns a goal-conditioned self-summary of the selected webpage. If
    \(a_t\in\mathcal{A}_{\mathrm{answer}}\), the trajectory terminates.

    \item \textbf{Reward $\mathcal{R}$:} During RL, each completed rollout
    \(\tau=(Q,\mathcal{I},\mathcal{H}_T,a_T)\) receives
    \begin{equation}
        r(\tau)
        = 0.5\,\mathbf{1}_{\mathrm{format}}(\tau)
        + \mathbf{1}_{\mathrm{answer}}(\tau).
        \label{eq:method-reward}
    \end{equation}
    Here \(\mathbf{1}_{\mathrm{format}}(\tau),
    \mathbf{1}_{\mathrm{answer}}(\tau)\in\{0,1\}\) indicate whether \(\tau\)
    satisfies the one-action protocol and produces a correct final answer,
    respectively. The format indicator is one only when every interaction turn
    is well formed. The answer indicator first uses exact matching and falls
    back to an LLM judge when exact matching is inconclusive.
\end{itemize}

\subsection{Factorized Adaptive Rollout}
\label{subsec:factorized_adaptive_rollout}

\begin{figure}[t]
    \centering
    \includegraphics[width=1.0\linewidth]{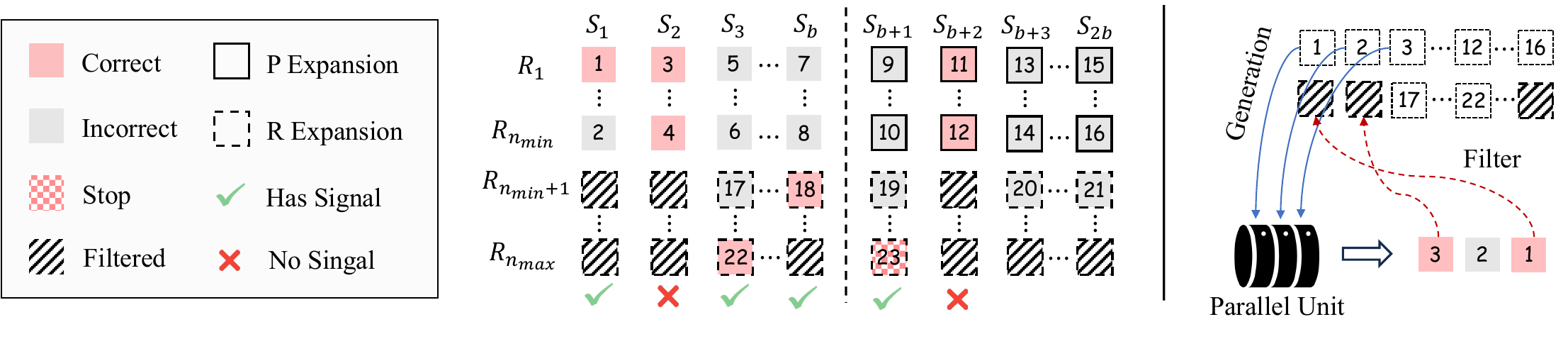}
    \caption{
    \textbf{FAR sampling details and filtering process.}
    The numbers 1, 2, 3, and subsequent indices indicate the intuitive sampling
    order. FAR fills base rollouts first, then allocates expanded rollout slots
    horizontally across groups. Once an answer-correct rollout appears for a
    group, later Rollout Allocation slots for that group are filtered, avoiding
    redundant rollouts.
    }
    \label{fig:far-detail}
    \vspace{4pt}
\end{figure}

\noindent\textbf{Budget factorization.}
FAR turns fixed-size rollout generation into
signal-aware budget allocation for multimodal agentic RL. As illustrated in
Fig.~\ref{fig:far}, it factorizes the budget into two directions: \textit{Prompt
Expansion} increases the candidate prompt-group pool, while \textit{Rollout
Allocation} allocates extra rollout slots horizontally across groups at the same
rollout depth. Fig.~\ref{fig:far-detail} gives the detailed order. FAR first
fills the base vertical rollouts for each candidate group, then advances
expanded slots by depth across groups, so hard groups receive additional
attempts while groups that already provide answer-level variation stop consuming
redundant tail slots.

\noindent\textbf{Signal evaluation.}
For rollout scheduling, FAR uses the answer indicator, rather than the total
scalar reward, to identify whether a prompt provides a relative training signal.
Let \(x\) denote a prompt and let
\(\mathcal{G}_K(x)=\{\tau_1,\ldots,\tau_K\}\) be its completed rollout group.
We mark each rollout by answer correctness and define useful answer-level
reward variation as mixed answer-correct and answer-incorrect outcomes:
\begin{equation}
    y_i=\mathbf{1}_{\mathrm{answer}}(\tau_i), \qquad
    \mathrm{Sig}(x;\mathcal{G}_K)
    = \mathbf{1}\left\{0 < \sum_{i=1}^{K} y_i < K\right\}.
    \label{eq:method-far-signal}
\end{equation}
Only groups satisfying Eq.~\ref{eq:method-far-signal} are retained for the
policy update. Groups with all correct or all incorrect final answers are
filtered from the update because they do not provide within-group answer
variation.

\noindent\textbf{Rollout filtering.}
The same signal definition controls which expanded slots remain active. Rollout
slots are defined upfront rather than generated by two separate stages. Each
candidate group has \(n_{\min}\) base slots and at most \(n_{\max}\) total
slots. After the first \(k\) observed rollouts of a group, the next expanded
slot remains active only if all observed rollouts are answer-incorrect:
\begin{equation}
    \alpha_k(x)
    =
    \mathbf{1}\left\{\sum_{i=1}^{k} y_i = 0\right\},
    \qquad
    n_{\min} \leq k < n_{\max},
    \label{eq:method-rollout-mask}
\end{equation}
where \(\alpha_k(x)=0\) masks later expanded slots for that group. This
``expand-only-all-wrong'' rule spends extra attempts on hard groups that still
lack an answer-correct trajectory, and filters redundant rollout tails as soon
as such a trajectory is observed. The horizontal direction in
Fig.~\ref{fig:far}(c) and the ordered indices in Fig.~\ref{fig:far-detail} show
that FAR expands rollouts breadth-first across eligible groups, rather than
spending the extra budget deeply on one group at a time. The rightmost panel of
Fig.~\ref{fig:far-detail} illustrates this filtering process: after the parallel
generation unit returns an answer-correct sample, subsequent allocation slots
from the same group are masked before they enter rollout generation.

\noindent\textbf{Stopping rule.}
A group is counted as valid only after it has completed at least \(n_{\min}\)
rollouts and satisfies Eq.~\ref{eq:method-far-signal}. FAR stops rollout
generation when either enough valid groups have been collected for the policy
update, or the accounted rollout slots reach a near-complete threshold. The
latter counts both completed slots and slots skipped by the rollout mask, and is
set to \(0.99\) of all pre-allocated slots by default. Because FAR creates
redundant candidate slots, this condition prevents a training step from waiting
for the last few slow or stalled tool-interleaved trajectories.

\noindent\textbf{Rollout selection.}
After rollout generation, FAR filters groups by
Eq.~\ref{eq:method-far-signal}. Since a retained group can complete more
rollouts than the update budget requires, FAR selects at most \(n_{\min}\)
trajectories as the update subset \(\mathcal{U}(x)\). The default strategy
preserves reward diversity and coarse behavior diversity by preferring
trajectories with different rewards, reasoning rounds, and expanded tool-call
counts. Non-selected rollouts are filtered from the gradient update but still
contribute to the full-group outcome estimate below. Let \(\mathcal{U}\) denote
the union of selected trajectories in a minibatch.

\noindent\textbf{Policy update.}
FAR uses an RLOO estimator for policy optimization. It computes the
leave-one-out baseline from the full completed group, so expanded rollouts can improve the
outcome estimate even when only a compact subset enters the update. Here
\(r(\cdot)\) denotes the reward defined in
Eq.~\ref{eq:method-reward}:
\begin{equation}
    \mu_x=\frac{1}{K}\sum_{j=1}^{K} r(\tau_j),\qquad
    \hat{A}_i
    =
    r(\tau_i)-\frac{1}{K-1}\sum_{j\neq i}r(\tau_j)
    =
    \frac{K}{K-1}(r(\tau_i)-\mu_x),
    \quad \tau_i\in\mathcal{U}(x).
    \label{eq:method-full-group-rloo}
\end{equation}
The scalar advantage is applied only to policy-generated tokens; tool-observation
tokens are masked. The policy is optimized with the clipped objective
\begin{equation}
\begin{aligned}
    \mathcal{J}(\theta)
    =
    \mathbb{E}\left[
    \frac{1}{|\mathcal{U}|}
    \sum_{\tau_i\in\mathcal{U}}
    \frac{1}{|\Omega_i|}
    \sum_{t\in \Omega_i}
    \min\left(
        \rho_{i,t}(\theta)\hat{A}_i,\,
        \mathrm{clip}(\rho_{i,t}(\theta),
        1-\epsilon_{\mathrm{low}},1+\epsilon_{\mathrm{high}})\hat{A}_i
    \right)
    \right],
\end{aligned}
\label{eq:method-clipped-objective}
\end{equation}
where \(\Omega_i\) denotes policy-generated token positions and \(\rho_{i,t}\) is
the token-level importance ratio. The lower and upper clipping ranges are
written separately to match the asymmetric clipping used in implementation.

\subsection{Evidence-Verified Reasoning}
\label{subsec:evidence_verified_reasoning}

\noindent\textbf{Evidence verification.}
SimpleSearch-VL makes retrieved evidence directly checkable inside the model
context. Text-search results provide candidate snippets, titles, and URLs for
page-level verification, while region-level reverse image search returns
thumbnails together with the associated title and URL, allowing the model to
compare the visual match against the queried region before using the source. As illustrated in the
middle panel of Fig.~\ref{fig:markov}, the agent is trained to follow this
explicit chain: inspect candidate evidence, verify whether it supports the
current hypothesis, and then choose whether to search, visit, or answer. During
SFT data construction, we preserve the original tool calls and observations but
rewrite assistant reasoning to expose these verification decisions; details are
given in Appendix~\ref{app:evidence-aware-data}. 

\noindent\textbf{Region-level cache.}
Completed tool responses are cached during both training and inference to
reduce repeated external calls. Text search and webpage visit use SHA-256 keys
over the normalized query and URL, respectively. Image search additionally
requires a region-aware rule because nearby predicted boxes over the same visual
target should share results. For an input image with content hash \(h\) and
queried box \(b\), the system searches cached boxes \(\mathcal{C}(h)\) and
reuses the nearest cached result when
\begin{equation}
    b^\star=\arg\max_{\tilde{b}\in\mathcal{C}(h)}
    \operatorname{IoU}(b,\tilde{b}),\qquad
    \operatorname{IoU}(b,b^\star)\ge \tau_{\mathrm{cache}},
\end{equation}
where \(\tau_{\mathrm{cache}}=0.7\) in our experiments. Otherwise, the tool
performs online reverse image search and appends the new box-result pair to the cache.
This keeps reverse image search region-specific while avoiding repeated calls for
slightly different boxes around the same visual evidence. In later training
stages, this strategy makes most repeated image-search requests cache hits,
reducing the additional image-search cost to nearly zero.

\subsection{Self-Summarized Visit}
\label{subsec:self_summarized_visit}

SimpleSearch-VL replaces the external webpage summarizer with self-summary by
the policy model itself during both training and evaluation. Given selected URLs
and an extraction goal, \texttt{visit} fetches and normalizes the webpages, and
the policy model produces a concise goal-conditioned summary that is appended to
the next Markov state in Eq.~\ref{eq:agent-state}. This avoids deploying an
additional summarization service and allows webpage-summary behavior to adapt
together with the agentic policy. Table~\ref{tab:self-summary-ablation} shows
that self-summary improves accuracy over external summarizers while adding only
moderate inference-time overhead.

\section{Experiments}
\label{sec:experiments}

\setlength{\textfloatsep}{2pt plus 1pt minus 1pt}
\setlength{\floatsep}{2pt plus 1pt minus 1pt}
\setlength{\intextsep}{2pt plus 1pt minus 1pt}
\setlength{\dbltextfloatsep}{2pt plus 1pt minus 1pt}
\setlength{\dblfloatsep}{2pt plus 1pt minus 1pt}

\subsection{Implementation Details}

\textbf{Training Data.}
Our supervised fine-tuning data are generated directly with the same agentic
search workflow used for RL and evaluation. Specifically,
\texttt{Qwen3-VL-235B-A22B-Instruct} autonomously produces multi-turn
tool-interleaved trajectories, turning the construction of complex search
trajectories into a scalable rollout process once the workflow is defined. We
then use \texttt{gemini-3.1-pro} to audit each trajectory and rewrite only the
assistant reasoning, filtering unsupported answers, low-value searches, and
weak evidence chains while preserving the original tool calls and observations.
The final training pool contains \textbf{5K} evidence-aware SFT trajectories,
and RL uses \textbf{2K} RL data from the same task distribution.
Appendix~\ref{app:evidence-aware-data} provides the detailed construction and
quality-control criteria.

\textbf{Benchmarks}
We evaluate on six benchmarks that stress different aspects of multimodal web search and visual knowledge reasoning: MMSearch~\citep{mmsearch}, MMSearch+~\citep{mmsearch}, BrowseComp-VL~\citep{webwatcher}, FVQA~\citep{wang2017fvqa}, LiveVQA~\citep{fu2025livevqa}, and SimpleVQA~\citep{cheng2025simplevqa}. Together, these benchmarks cover visual entity recognition, open-world information seeking, evidence-grounded web retrieval, multi-hop reasoning, and long-tail VQA.

\textbf{Tool Definitions.}
SimpleSearch-VL uses a minimal tool interface for multimodal evidence
acquisition: text search, region-level reverse image search, and webpage visit with
goal-conditioned self-summary. Complete tool declarations, including each
tool's input signature and returned evidence fields, are provided in
Appendix~\ref{app:agent-workflow}, Table~\ref{tab:search-tools}.

\textbf{Training Details.}
SimpleSearch-VL is instantiated with \textit{Qwen3-VL-8B-Instruct} and
\textit{Qwen3-VL-30B-A3B-Instruct}~\citep{Qwen3-VL}, using
ms-swift~\citep{ms_swift} for agentic SFT and rLLM~\citep{rllm2025} for
agentic RL. On 8 H200 GPUs, the 8B model takes about 2 hours for SFT and 16
hours for RL; the 30B-A3B model takes about 3 hours for SFT and 24 hours for RL. More
training configurations are provided in Appendix~\ref{app:implementation-details}.

\subsection{Main Results}
\label{subsec:main-results}

\begin{table}[!t]
    \setlength{\abovecaptionskip}{2pt}
    \setlength{\belowcaptionskip}{2pt}
    \vspace{5pt}
    \caption{
    Main results on multimodal agentic search benchmarks.
    \textbf{Bold} and \underline{underline} mark the best and second-best reported score in each column.
    ``--'' denotes unreported results.
    \textsuperscript{\dag} indicates random-subset evaluation rather than the full evaluation split. Literature results follow their source protocols; LMM-Searcher and ODE use single-image MMSearch+.
    }
    \centering
    \scriptsize
    \setlength{\tabcolsep}{3.2pt}
    \resizebox{\textwidth}{!}{
    \begin{tabular}{l cccccc}
    \toprule
    \textbf{Model} & \textbf{MMSearch} & \textbf{MMSearch+} & \textbf{BC-VL} & \textbf{FVQA} & \textbf{LiveVQA} & \textbf{SimpleVQA} \\
    \midrule
    \multicolumn{7}{c}{\textbf{Direct Answer}} \\
    Gemini-3-Pro~\citep{google2025gemini3pro} & 65.9 & 26.4 & 41.4 & 58.9 & 51.1 & -- \\
    Claude-Opus-4.6~\citep{Claude_46} & 59.8 & 13.2 & 43.5 & 60.1 & 53.1 & -- \\
    Kimi-K2.5~\citep{kimiteam2026kimik25visualagentic} & 65.6 & 9.7 & 27.6 & 59.6 & 57.3 & -- \\
    Qwen3-VL-8B~\citep{Qwen3-VL} & 15.2 & 3.2 & 25.1 & 28.0 & 41.0 & 42.9 \\
    Qwen3-VL-30B-A3B~\citep{Qwen3-VL} & 18.7 & 3.2 & 29.6 & 34.7 & 42.7 & 53.2 \\
    \midrule
    \multicolumn{7}{c}{\textbf{Agentic Workflow}} \\
    Gemini-3-Pro~\citep{google2025gemini3pro} & \underline{82.9} & 33.5 & 51.8 & 76.7 & \underline{79.9} & -- \\
    Claude-Opus-4.6~\citep{Claude_46} & 76.2 & 31.3 & 48.3 & 74.5 & 67.4 & -- \\
    Kimi-K2.5~\citep{kimiteam2026kimik25visualagentic} & 76.6 & 27.8 & 50.3 & 76.5 & 76.6 & -- \\
    Qwen3-VL-8B~\citep{Qwen3-VL} & 62.0 & 13.5 & 36.6 & 65.3 & 54.6 & 63.4 \\
    Qwen3-VL-30B-A3B~\citep{Qwen3-VL} & 63.2 & 14.1 & 44.6 & 67.4 & 61.9 & 66.6\\
    \midrule
    \multicolumn{7}{c}{\textbf{Multimodal Search Agents}} \\
    MMSearch-R1-7B~\citep{mmsearch-r1} & 53.8 & -- & 19.1 & 58.4 & 48.4 & 57.4 \\
    SVRL-full-7B~\citep{sathish2026svrl} & 60.3 & -- & -- & 65.3 & 57.8 & 58.3 \\
    ProMMSearchAgent-7B~\citep{yan2026prommsearchagent} & 55.2 & -- & -- & 63.1 & 32.5 & 58.6 \\
    DeepEyesV2-7B~\citep{deepeyesv2} & 63.7 & 9.5 & 24.8 & 60.6 & 58.0 & 59.4 \\
    SenseNova-MARS-8B~\citep{chng2025sensenova} & 67.8 & -- & -- & 67.1 & 56.2 & 70.2 \\
    MM-DeepResearch-8B~\citep{yao2026mm} & 67.8 & -- & 37.9 & 69.2 & 65.0 & 65.9 \\
    Vision-DeepResearch-8B~\citep{huang2026vision} & 69.6 & 20.4 & 42.6 & 64.7\textsuperscript{\dag} & 76.7\textsuperscript{\dag} & -- \\
    POINTS-Seeker-8B~\citep{liu2026points} & 70.8 & 25.2 & 44.4 & 71.2 & 77.7 & 68.8 \\
    OpenSearch-VL-8B~\citep{chen2026opensearch} & 64.5 & -- & 37.6 & 71.5 & 59.6 & 71.6 \\
    Visual-Seeker-8B~\citep{zhang2026visualseeker} & 72.2 & 27.3 & 47.6 & -- & -- & -- \\
    ODE-8B~\citep{huang2026onpolicy} & 66.0 & 24.9 & 41.9 & 64.7\textsuperscript{\dag} & -- & 70.3\textsuperscript{\dag} \\
    \rowcolor[HTML]{F3F5F8}
    \textbf{SimpleSearch-VL-8B (ours)} & 77.1 & 32.5 & 52.1 & \underline{76.8} & 75.2 & \underline{76.6} \\
    \midrule
    WebWatcher-32B~\citep{webwatcher} & 55.3 & 11.5 & 27.0 & 64.3 & 58.7 & 59.0 \\
    SenseNova-MARS-32B~\citep{chng2025sensenova} & 74.3 & -- & -- & 72.6 & 60.8 & 74.1 \\
    Vision-DeepResearch-30B-A3B~\citep{huang2026vision} & 69.6 & 28.5 & 53.7 & 74.2\textsuperscript{\dag} & 77.6\textsuperscript{\dag} & -- \\
    OpenSearch-VL-30B-A3B~\citep{chen2026opensearch} & 68.7 & -- & 41.1 & 73.2 & 67.4 & 74.9 \\
    RedSearcher-30B-A3B~\citep{chu2026redsearcher} & 72.9 & 26.6 & \textbf{57.2} & -- & {79.3} & -- \\
    LMM-Searcher-30B (100 turns)~\citep{du2026towards} & 72.3 & \textbf{34.8} & -- & -- & -- & -- \\
    \rowcolor[HTML]{F3F5F8}
    \textbf{SimpleSearch-VL-30B-A3B (ours)} & \textbf{83.6} & \underline{34.4} & \underline{55.9} & \textbf{79.0} & \textbf{81.1} & \textbf{79.6}  \\
    \bottomrule
    \end{tabular}
    }
    \label{tab:search-qa-main}
    \vspace{0.6cm}
\end{table}

Table~\ref{tab:search-qa-main} reports the main evaluation results. Compared
with the corresponding untuned Qwen3-VL~\citep{Qwen3-VL} agentic baselines,
SimpleSearch-VL improves the average score by \textbf{15.8} points for the 8B
model and \textbf{16.0} points for the 30B-A3B model on the shared evaluation
benchmarks. Notably, SimpleSearch-VL-8B also outperforms the larger
OpenSearch-VL-30B-A3B~\citep{chen2026opensearch} on all five shared
benchmarks, with gains of 8.4 points on MMSearch, 11.0 on BrowseComp-VL, 3.6
on FVQA, 7.8 on LiveVQA, and 1.7 on SimpleVQA. This cross-scale comparison
suggests that the training and tool-use design can be more important than
increasing model size alone.

SimpleSearch-VL-30B-A3B is also competitive with stronger proprietary agentic
systems. Compared with agentic Gemini-3-Pro~\citep{comanici2025gemini}, it is
higher on five benchmarks. These results are obtained with 5K SFT
trajectories and 2K RL data, compared with the 36K SFT trajectories and 8K RL
data used by OpenSearch-VL; as summarized in
Table~\ref{tab:training-cost-comparison}, the representative average increases
from 58.3 to 70.3 for the 8B model and from 62.6 to 74.9 for the 30B-A3B
model.

\subsection{Ablation Studies}
\label{subsec:ablation}

Unless otherwise specified, ablations are conducted on Qwen3-VL-8B-Instruct~\citep{Qwen3-VL} under the direct-RL setting: the model is trained for 100 steps on 914 FVQA-train~\citep{wang2017fvqa} samples with a batch size of 64 and $n=4$ rollouts. For evaluation, all benchmarks except MMSearch~\citep{mmsearch} use a sampled subset of 100 examples. In the rollout analysis, \textit{signal rate} is the fraction of prompt groups with mixed correct and incorrect rollouts.

\subsubsection{Rollout Infrastructure Analysis.}
Rather than directly comparing with a specific rollout method, we conduct a systematic analysis by decomposing rollout infrastructure into two forms of budget expansion. The first is \textit{Prompt Expansion}, which expands the number of prompts sampled for rollout generation and is the main optimization target of recent methods. For example, DAPO-style strategies in SLIME~\citep{slime_github} keep expanding prompts until the required number of effective samples is obtained. We do not adopt this setting, as it would make time-cost comparisons unfair. Instead, we expand prompts by fixed multipliers and stop once the termination condition is reached. The second is \textit{Rollout Allocation}, which expands the number of rollouts within a prompt group when additional samples may still produce useful reward contrast. In implementation, we replace the original sampling strategy with an interleaved scheme and introduce a signal-ratio threshold: once a correct response appears for a sample, its subsequent expanded rollouts are dynamically masked out. Finally, we analyze our proposed \textbf{FAR} strategy, which combines \textit{Prompt Expansion} and \textit{Rollout Allocation}, while further improving the sampling order and termination criteria.

\begin{figure}[H]
    \centering
    \begin{subfigure}[t]{0.29\textwidth}
        \vspace{0pt}
        \centering
        \begin{minipage}[t][3.05cm][c]{\linewidth}
            \centering
            \small
            \setlength{\tabcolsep}{3pt}
            \resizebox{\linewidth}{!}{%
            \begin{tabular}{lcccc}
                \toprule
                \textbf{Prompt} & \textbf{AVG} & \textbf{MMS} & \textbf{BC-VL} & \textbf{FVQA} \\
                \midrule
                $1\times$   & 57.8 & 70.3 & 33.0 & 70.0 \\
                $1.5\times$ & 59.5 & 69.6 & 37.0 & 72.0 \\
                $2\times$   & 62.7 & 72.2 & 43.0 & 73.0 \\
                $3\times$   & 62.5 & 73.6 & 43.0 & 71.0 \\
                \bottomrule
            \end{tabular}
            }
        \end{minipage}
        \vspace{-10pt}
        \caption{Prompt-expansion scores.}
        \label{fig:prompt-expansion-metrics}
    \end{subfigure}
    \setcounter{subfigure}{1}
    \begin{subfigure}[t]{0.35\textwidth}
        \vspace{0pt}
        \centering
        \begin{minipage}[t][3.05cm][c]{\linewidth}
            \centering
            \includegraphics[width=\linewidth]{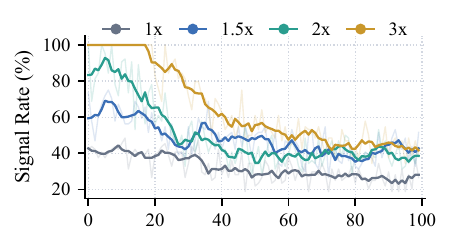}
        \end{minipage}
        \vspace{-10pt}
        \caption{Signal rate.}
        \label{fig:prompt-expansion-signal}
    \end{subfigure}
    \begin{subfigure}[t]{0.34\textwidth}
        \vspace{0pt}
        \centering
        \begin{minipage}[t][3.05cm][c]{\linewidth}
            \centering
            \includegraphics[width=\linewidth]{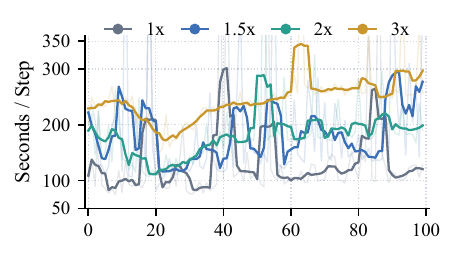}
        \end{minipage}
        \vspace{-10pt}
        \caption{Rollout time.}
        \label{fig:prompt-expansion-time}
    \end{subfigure}
    \vspace{-3pt}
    \caption{\textbf{Prompt expansion trade-off.} (a) Best evaluation scores under different expansion ratios. (b) Signal rate during training. (c) Per-step rollout time.}
    \label{fig:prompt-expansion}
    \vspace{10pt}
\end{figure}

\noindent\textbf{Prompt Expansion.}
We vary the prompt expansion ratio while stopping once the target number of
valid groups is reached, and compare its effect on accuracy, signal
availability, and rollout time. As shown in
Fig.~\ref{fig:prompt-expansion-metrics}, Prompt Expansion improves performance
by adding more effective samples within each training step, thereby
accelerating convergence. Fig.~\ref{fig:prompt-expansion-signal} supports this
trend: larger expansion ratios provide more signals in the early stage, but the
gain becomes marginal later because most prompt groups become fully correct and
no longer provide useful reward variation.
Fig.~\ref{fig:prompt-expansion-time} shows that rollout time increases with the
expansion ratio. Considering both accuracy and time cost, $2\times$ provides a reasonable trade-off.

\begin{figure}[H]
    \centering
    \begin{subfigure}[t]{0.29\textwidth}
        \vspace{0pt}
        \centering
        \begin{minipage}[t][3.05cm][c]{\linewidth}
        \centering
        \small
        \renewcommand{\arraystretch}{1.1}
        \setlength{\tabcolsep}{2.5pt}
        \resizebox{\linewidth}{!}{%
        \begin{tabular}{lccccc}
            \toprule
            \textbf{Rollouts}  & \textbf{AVG} & \textbf{MMS} & \textbf{BC-VL} & \textbf{FVQA} \\
            \midrule
            $n=2$   & 56.4 & 66.1 & 34.0 & 69.0 \\
            $n=4$   & 58.1 & 67.3 & 40.0 & 67.0 \\
            $n=6$   & 62.6 & 71.9 & 47.0 & 69.0 \\
            $n=8$   & 61.0 & 69.0 & 46.0 & 68.0 \\
            $n=10$  & 63.1 & 71.3 & 48.0 & 70.0 \\
            \bottomrule
        \end{tabular}
        }
        \end{minipage}
        \vspace{-10pt}
        \caption{Rollout-budget scores.}
        \label{fig:rollout-allocation-metrics}
    \end{subfigure}
    \begin{subfigure}[t]{0.34\textwidth}
        \vspace{0pt}
        \centering
        \begin{minipage}[t][3.05cm][c]{\linewidth}
            \centering
            \includegraphics[width=\linewidth]{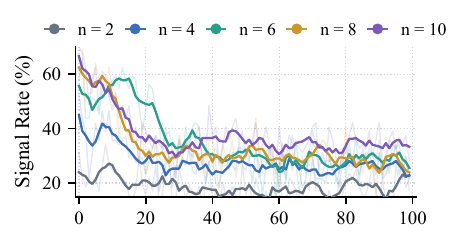}
        \end{minipage}
        \vspace{-10pt}
        \caption{Signal rate.}
        \label{fig:rollout_signal}
    \end{subfigure}
    \begin{subfigure}[t]{0.34\textwidth}
        \vspace{0pt}
        \centering
        \begin{minipage}[t][3.05cm][c]{\linewidth}
            \centering
            \includegraphics[width=\linewidth]{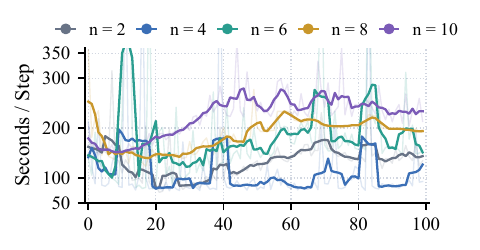}
        \end{minipage}
        \vspace{-10pt}
        \caption{Rollout time.}
        \label{fig:rollout_time}
    \end{subfigure}
    \vspace{-3pt}
    \caption{\textbf{Fixed rollout-budget trade-off.} (a) Best evaluation scores for each rollout budget. (b) Signal rate during training. (c) Per-step rollout time.}
    \label{fig:rollout-allocation}
    \vspace{10pt}
\end{figure}

\begin{figure}[H]
    \centering
    \begin{subfigure}[t]{0.30\textwidth}
        \vspace{0pt}
        \centering
        \begin{minipage}[t][3.05cm][c]{\linewidth}
            \centering
            \small
            \renewcommand{\arraystretch}{1.28}
            \setlength{\tabcolsep}{3pt}
            \resizebox{\linewidth}{!}{%
            \begin{tabular}{lcc}
                \toprule
                \textbf{Method} & \textbf{Recovered Groups} & \textbf{Rate} \\
                \midrule
                Rollout 4--6 & 117 & 11.7\% \\
                Rollout 4--8 & 195 & 17.7\% \\
                \bottomrule
            \end{tabular}
            }
        \end{minipage}
        \vspace{-14pt}
        \caption{Recovery metrics.}
        \label{fig:adaptive-recovery-metrics}
    \end{subfigure}
    \begin{subfigure}[t]{0.34\textwidth}
        \vspace{0pt}
        \centering
        \begin{minipage}[t][3.05cm][c]{\linewidth}
            \centering
            \includegraphics[width=\linewidth]{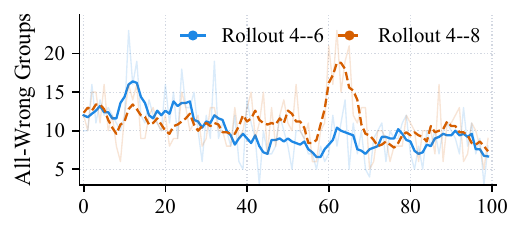}
        \end{minipage}
        \vspace{-14pt}
        \caption{All-wrong groups.}
        \label{fig:adaptive-recovery-all-wrong}
    \end{subfigure}
    \begin{subfigure}[t]{0.34\textwidth}
        \vspace{0pt}
        \centering
        \begin{minipage}[t][3.05cm][c]{\linewidth}
            \centering
            \includegraphics[width=\linewidth]{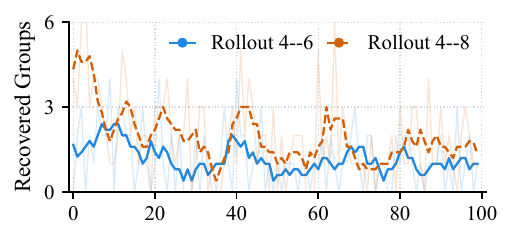}
        \end{minipage}
        \vspace{-14pt}
        \caption{Recovered groups.}
        \label{fig:adaptive-recovery-recovered}
    \end{subfigure}
    \vspace{-3pt}
    \caption{\textbf{Hard-group recovery with additional rollouts.} (a) Recovered occurrences and recovery rate among hard groups. (b) Initially all-wrong group occurrences per step. (c) Occurrences recovered by later rollouts.}
    \label{fig:adaptive_recovery}
    \vspace{10pt}
\end{figure}

\noindent\textbf{Rollout Allocation.}
We analyze Rollout Allocation from two perspectives. \textit{First, how does increasing the base rollout number $n$ affect performance and efficiency?} As shown in Fig.~\ref{fig:rollout_time}, larger $n$ leads to higher per-step time cost, while generally improving performance. However, Fig.~\ref{fig:rollout_signal} shows that different values of $n$ exhibit clear signal differences at the early stage of training, but the gap becomes much smaller later. This trend is intuitive: as the model becomes stronger, the proportion of fully correct samples increases, leaving fewer samples with meaningful rollout-level variation. \textit{Second, how efficiently can Rollout Allocation recover useful training samples during training?} As shown in Fig.~\ref{fig:adaptive_recovery}, the 4--8 setting recovers more hard group occurrences than the 4--6 setting. We compute the recovery rate over training-step group occurrences whose base rollouts are all incorrect, counting repeated prompt occurrences across steps separately. The recovery rate increases from 11.7\% to 17.7\%, indicating that under the 4--8 setting, about \textbf{17.7\%} of these hard occurrences can produce at least one correct later rollout and thus participate in training. This is a favorable outcome: Rollout Allocation not only improves training efficiency, but also makes better use of hard samples.

\begin{figure}[H]
    \centering
    \begin{subfigure}[t]{0.32\textwidth}
        \vspace{0pt}
        \centering
        \begin{minipage}[t][3.05cm][c]{\linewidth}
            \centering
            \small
            \renewcommand{\arraystretch}{1.12}
            \setlength{\tabcolsep}{2.5pt}
            \resizebox{\linewidth}{!}{%
            \begin{tabular}{lccccc}
                \toprule
                \textbf{Method} & \textbf{Step(s)}& \textbf{AVG}& \textbf{MMS} & \textbf{BC-VL} & \textbf{FVQA}  \\
                \midrule
                Standard & 192  & 56.6 & 66.8 & 35.0 & 68.0   \\
                Rollout 4--6 & 118 & 56.8 & 65.4 & 37.0 & 68.0  \\
                Rollout 4--8 & 134 & 60.4 & 67.2 & 41.0 & 73.0  \\
                Prompt 2$\times$ & 273 & 59.7 & 67.2 & 43.0 & 69.0   \\
                \midrule
                FAR (2$\times$,4--6) & 165 & 62.0 & 73.0  & 45.0 & 68.0 \\
                FAR (2$\times$,4--8) & 187 & 62.8 & 71.3 & 45.0 & 72.0 \\
                \bottomrule
            \end{tabular}
            }
        \end{minipage}
        \vspace{-10pt}
        \caption{FAR scores.}
        \label{fig:far-ablation-metrics}
    \end{subfigure}
    \begin{subfigure}[t]{0.33\textwidth}
        \vspace{0pt}
        \centering
        \begin{minipage}[t][3.05cm][c]{\linewidth}
            \centering
            \includegraphics[width=\linewidth]{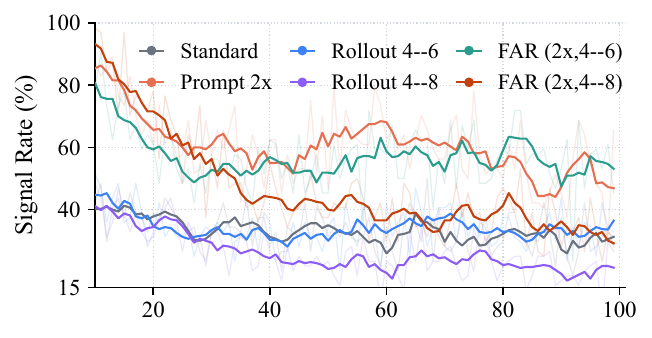}
        \end{minipage}
        \vspace{-10pt}
        \caption{Signal rate.}
        \label{fig:far-ablation-signal}
    \end{subfigure}
    \begin{subfigure}[t]{0.33\textwidth}
        \vspace{0pt}
        \centering
        \begin{minipage}[t][3.05cm][c]{\linewidth}
            \centering
            \includegraphics[width=\linewidth]{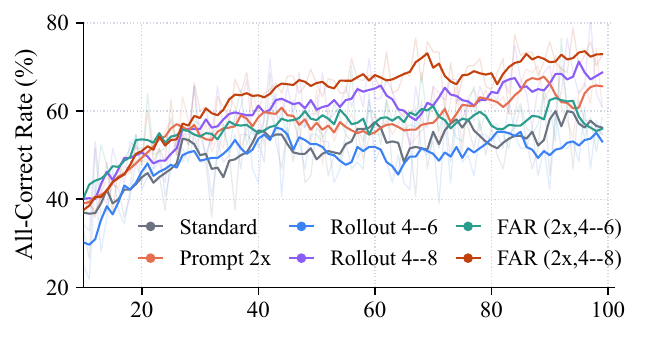}
        \end{minipage}
        \vspace{-10pt}
        \caption{All-correct rate.}
        \label{fig:far-ablation-all-correct}
    \end{subfigure}
    \vspace{-5pt}
    \caption{\textbf{FAR accuracy--efficiency trade-off.} (a) Rollout cost and evaluation scores. (b) Signal rate. (c) All-correct rate.}
    \label{fig:far-ablation}
    \vspace{10pt}
\end{figure}

\noindent\textbf{Factorized Adaptive Rollout.}
Fig.~\ref{fig:far-ablation-metrics} shows that expanding the rollout budget through Rollout Allocation brings clear performance gains. We do not emphasize the reward curve in this setting because the average reward becomes less comparable under adaptive sampling: Rollout Allocation continues sampling harder prompt groups, so a lower observed reward can reflect a harder sampled distribution rather than weaker model behavior. We therefore analyze the training signal rate instead. As shown in Fig.~\ref{fig:far-ablation-signal}, both increasing rollout allocation and FAR allocation budget from 6 to 8 lead to a faster decline in signal rate. This indicates accelerated convergence, as more prompt groups become fully correct and no longer provide useful within-group reward variance; Fig.~\ref{fig:far-ablation-all-correct} shows this trend more directly through the rising all-correct rate. In terms of efficiency, FAR (2$\times$, 4--8) achieves a rollout time comparable to the Standard setting, while improving the average score by \textbf{6.2} points.

\subsubsection{Visual Evidence Verification}

Table~\ref{tab:evidence-aware-ablation} compares direct training on trajectories distilled from the Qwen3-VL-235B-A22B model with training on trajectories augmented by textual and visual evidence-aware reasoning. Under the SFT-only setting, evidence-aware training improves the average score by \textbf{3.4} points. Applying RL on top of the evidence-aware SFT model further improves the average score by \textbf{5.4} points, suggesting that explicit evidence verification provides a stronger initialization for policy optimization. We further remove visual thumbnails from the image-search observations during evaluation. This consistently causes an average drop of about three points for both SFT and RL models, showing that thumbnails are not merely auxiliary metadata: \textit{they provide the visual cue needed to verify whether a reverse-image-search result matches the queried entity.} Qualitative cases, including Fig.~\ref{fig:markov}, show that without thumbnail-based verification, the model can incorrectly rely on visually mismatched image-search results. This verification signal introduces negligible overhead, since each thumbnail typically contains fewer than 0.1M pixels, which is minor compared with the original high-resolution image input.

\begin{table}[H]
    \centering
    \vspace{5pt}
    \caption{\textbf{Evidence verification ablation.} ``Direct'' refers to training using the distillation trajectory directly. ``w/o thumbnail'' removes image-search thumbnails from the agent loop during inference.}
    \vspace{-3pt}
    \label{tab:evidence-aware-ablation}
    \scriptsize
    \renewcommand{\arraystretch}{1.18}
    \setlength{\tabcolsep}{5pt}
    \resizebox{0.65\linewidth}{!}{
    \begin{tabular}{lccccc}
        \toprule
        \textbf{Method} & \textbf{AVG} & \textbf{MMSearch} & \textbf{BC-VL}& \textbf{FVQA} & \textbf{LiveVQA} \\
        \midrule
        Direct SFT & 61.0 & 71.9 & 36.0 & 72.0 & 64.0 \\
        Direct RL & 64.9\gain{3.9} & 73.6\gain{1.7} & 46.0\gain{10.0} & 71.0\drop{1.0} & 69.0\gain{5.0} \\
        \midrule
        Evidence-Aware SFT & 64.4\gain{3.4} & 74.4\gain{2.5} & 48.0\gain{12.0} & 70.0\drop{2.0} & 65.0\gain{1.0} \\
        \rowcolor{gray!8} w/o thumbnail & 61.9\drop{2.5} & 69.7\drop{4.7} & 44.0\drop{4.0} & 70.0\nochg & 64.0\drop{1.0} \\
        Evidence-Aware RL & 69.8\gain{5.4} & 77.2\gain{2.8} & 59.0\gain{11.0} & 76.0\gain{6.0} & 67.0\gain{2.0} \\
        \rowcolor{gray!8} w/o thumbnail & 66.1\drop{3.7} & 71.3\drop{5.9} & 54.0\drop{5.0} & 74.0\drop{2.0} & 65.0\drop{2.0} \\
        \bottomrule
    \end{tabular}
    }
    \vspace{-2mm}
\end{table}

\subsubsection{Self-Summarized Visit}

Self-summarized visit removes the need to deploy a separate summarization model for webpage reading. This matters for deep-search agents because webpages can contain tens of thousands of tokens, making external summarization a substantial training-time bottleneck at scale. It also increases inference deployment cost, since the agent's performance would depend on a second model rather than on the agent itself. We therefore use self-summary during both training and inference. Table~\ref{tab:self-summary-ablation} shows two main findings. \textbf{(1) Better agent-adapted summaries.} Self-summary outperforms external summarizers by about 1.8--2.6 average points, suggesting that training with self-summary can adapt summary generation to the information needs of the current agentic policy. \textbf{(2) Deployment-aware efficiency.} With the same two-node budget, the \(2{+}0\) self-summary setting is 28.5\% faster than the \(1{+}1\) external 8B summarizer setting, while also improving AVG by 1.8 points.

\begin{table}[H]
    \centering
    \vspace{5pt}
    \caption{\textbf{Self-summary versus external summarization.} Evaluation scores and inference time under different serving layouts. ``A+S'' denotes agent nodes plus external-summary nodes; self-summary uses no separate summary service.}
    \vspace{-3pt}
    \label{tab:self-summary-ablation}
    \scriptsize
    \renewcommand{\arraystretch}{1.22}
    \setlength{\tabcolsep}{3pt}
    \resizebox{0.9\linewidth}{!}{%
    \begin{tabular}{llccccccc}
        \toprule
        \textbf{Summary Source} & \textbf{Summary Model} & \textbf{Serving (A+S)} & \textbf{Time} & \textbf{AVG} & \textbf{MMSearch} & \textbf{BC-VL} & \textbf{FVQA} & \textbf{LiveVQA} \\
        \midrule
        Self-summary & Same 8B policy & 1+0 & 609s & 69.8 & 77.2 & 59.0 & 76.0 & 67.0 \\
        Self-summary & Same 8B policy & 2+0 & 358s & 69.7 & 77.8 & 58.0 & 76.0 & 67.0 \\
        \midrule
        External summarizer & Qwen3-VL-8B & 1+1 & 501s & 67.9\drop{1.9} & 76.6 & 56.0 & 71.0 & 68.0 \\
        External summarizer & Qwen3-VL-30B-A3B & 1+1 & 582s& 67.2\drop{2.6} & 77.8 & 53.0 & 71.0 & 67.0 \\
        \bottomrule
    \end{tabular}
    }
\end{table}

\subsubsection{Harness Comparison.}
We further compare different agentic harnesses under a zero-shot setting, where each harness is directly paired with an untuned 8B model. This isolates the effectiveness of the inference-time tool environment from post-training. As shown in Table~\ref{tab:harness-comparison}, SimpleSearch-VL improves the average score over Vision-DeepResearch by \textbf{2.7} points, validating the benefit of our decoupled tool design. Unlike Vision-DeepResearch, which couples reverse image search, webpage visit, and summary, or SenseNova-MARS, which couples search and visit, our harness lets the model independently plan search queries, select webpages to visit, and choose target entities for reverse image search. OpenSearch-VL adopts a similarly decoupled design, but our harness still improves the average score by \textbf{11.7} points.

\begin{table}[H]
    \centering
    \vspace{5pt}
    \caption{\textbf{Zero-shot harness comparison.} AVG is computed over common reported benchmarks; ``--'' denotes unavailable results. \textsuperscript{\dag} marks Vision-DeepResearch results on a random 300-sample subset.}
    \vspace{-3pt}
    \label{tab:harness-comparison}
    \scriptsize
    \renewcommand{\arraystretch}{1.16}
    \setlength{\tabcolsep}{4pt}
    \resizebox{0.80\linewidth}{!}{%
    \begin{tabular}{l l ccc c}
        \toprule
        \textbf{Harness} & \textbf{Model} & \textbf{MMSearch} & \textbf{FVQA} & \textbf{LiveVQA} & \textbf{AVG} \\
        \midrule
        SenseNova-MARS~\citep{chng2025sensenova} & Qwen3-VL-8B-Instruct & 47.4 & 53.6 & 39.4 & 46.8 \\
        OpenSearch-VL~\citep{chen2026opensearch} & Qwen3-VL-8B-Instruct & 37.4 & 58.7 & 50.6 & 48.9 \\
        Vision-DeepResearch~\citep{vdr} & Qwen3-VL-8B-Instruct & 52.0 & 58.7\textsuperscript{\dag} & 63.0\textsuperscript{\dag} & 57.9 \\
        MTA-DeepSearch~\citep{peng2026mta} & Qwen3-VL-8B-Instruct & 57.1 & 64.2 & -- & -- \\
        \midrule
        \rowcolor[HTML]{F3F5F8}
        \textbf{SimpleSearch-VL} & Qwen3-VL-8B-Instruct & 62.0 & 65.3 & 54.6 & 60.6 \\
        \bottomrule
    \end{tabular}
    }
\end{table}

\subsubsection{Round/Tool Usage Analysis}

\begin{figure}[H]
    \centering
    \begin{subfigure}[t]{0.28\textwidth}
        \vspace{7pt}
        \centering
        \begin{minipage}[t][3.05cm][c]{\linewidth}
            \centering
            \scriptsize
            \setlength{\tabcolsep}{2.5pt}
            \renewcommand{\arraystretch}{1.12}
            \resizebox{\linewidth}{!}{%
            \begin{tabular}{lccc}
                \toprule
                \textbf{Method} & \textbf{MMS} & \textbf{BC-VL} & \textbf{FVQA} \\
                \midrule
                SFT & 62.0 & 43.0 & 69.0 \\
                SFT+RL$_{step=10}$ & 61.4 & 47.0 & 68.0 \\
                SFT+RL$_{step=100}$ & 66.1 & 60.0 & 74.0 \\
                \midrule
                Direct RL$_{step=10}$ & 59.1 & 33.0 & 74.0 \\
                Direct RL$_{step=100}$ & 84.2 & 74.0 & 93.0 \\
                \bottomrule
            \end{tabular}
            }
        \end{minipage}
        \vspace{-2pt}
        \caption{Image-search ratio.}
        \label{fig:train-image-search-ratio}
    \end{subfigure}
    \hspace{0.02\linewidth}
    \begin{subfigure}[t]{0.68\textwidth}
        \vspace{7pt}
        \centering
        \begin{minipage}[t][3.05cm][c]{\linewidth}
            \centering
            \includegraphics[width=\linewidth]{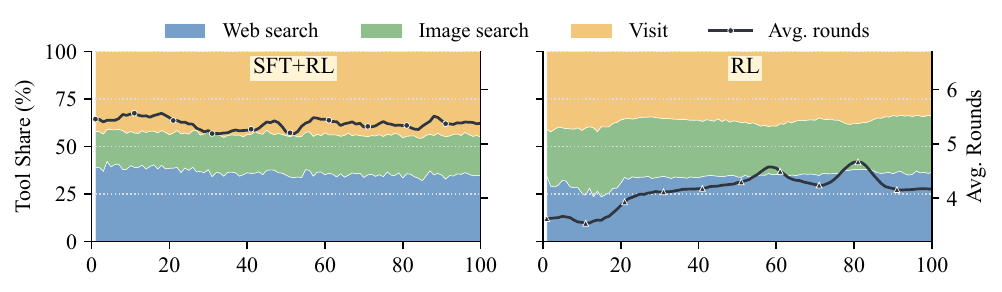}
        \end{minipage}
        \vspace{-2pt}
        \caption{Tool composition.}
        \label{fig:train-tool-composition}
    \end{subfigure}
    \vspace{-1mm}

    \caption{\textbf{Training-time tool behavior.} Left: image-search usage ratio. Right: tool-call composition and average reasoning rounds during training.}
    \label{fig:train-round-tool-usage}
    \vspace{10pt}
\end{figure}

Fig.~\ref{fig:train-round-tool-usage} compares SFT+RL and Direct RL under the same FAR (2$\times$, 4--6) setting. The SFT-initialized run keeps a longer and more stable search process, with average reasoning rounds changing only from 5.6 to 5.4 between steps 10 and 100. Direct RL starts from shorter trajectories and increases from 3.4 to 4.2 rounds, indicating that it is still learning when to search and verify evidence. The sample-level image-search ratio in Fig.~\ref{fig:train-image-search-ratio} shows a complementary pattern. Direct RL increases image-search invocation on all three reported benchmarks, from 59.1\% to 84.2\% on MMS, 33.0\% to 74.0\% on BC-VL, and 74.0\% to 93.0\% on FVQA. SFT+RL changes more moderately, from 61.4\% to 66.1\%, 47.0\% to 60.0\%, and 68.0\% to 74.0\% on the same benchmarks. This suggests that SFT provides a stronger tool-use prior, whereas Direct RL more readily expands reverse image search as a broadly useful evidence-gathering action.

\begin{figure}[H]
    \centering
    \includegraphics[width=0.80\linewidth]{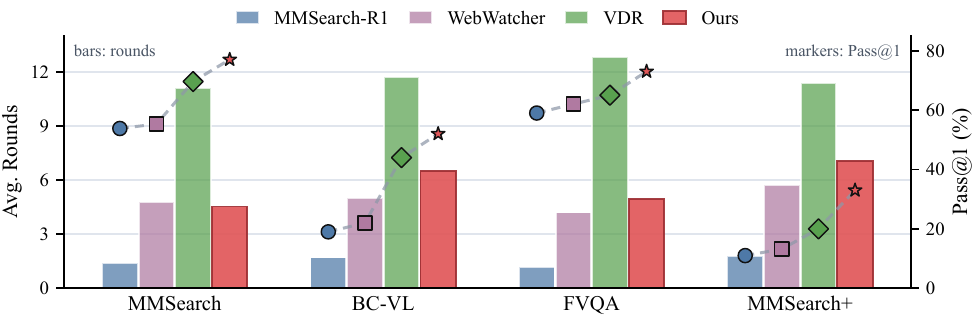}
    \vspace{-1mm}
    \caption{\textbf{Inference rounds and Pass@1.} Bars show the average number of reasoning rounds on each benchmark, while markers and dashed segments report the corresponding Pass@1 scores.}
    \label{fig:eval-round-tool-usage}
    \vspace{10pt}
\end{figure}

Fig.~\ref{fig:eval-round-tool-usage} reports inference rounds and Pass@1 across benchmarks. SimpleSearch-VL uses a moderate number of rounds overall, but adapts its reasoning depth to task difficulty: it uses about seven rounds on the harder MMSearch+ benchmark and about four rounds on MMSearch. This indicates that the trained agent does not follow a fixed search template, but adjusts its search depth to the task.

\subsubsection{Training Cost Comparison}

Table~\ref{tab:training-cost-comparison} shows that SimpleSearch-VL reaches strong multimodal search performance with substantially smaller training data and a much lighter training budget. Compared with OpenSearch-VL, our recipe uses about $7\times$ fewer SFT trajectories and $4\times$ less RL data. The compute difference is also pronounced: the 8B model requires only one H200 node for roughly 18 hours across SFT and RL, whereas OpenSearch-VL reports hundreds of H20 GPUs for multi-day SFT and RL stages. Despite this smaller scale, SimpleSearch-VL improves the representative average Pass@1 from 58.3 to 70.3 for the 8B model and from 62.6 to 74.9 for the 30B-A3B model. This suggests that the gains come not from scaling data or cluster size, but from improving the efficiency of rollout allocation, making visual evidence explicitly verifiable, and keeping webpage summarization inside the agent itself.

\begin{table}[H]
    \centering
    \vspace{5pt}
    \caption{\textbf{Training data and compute efficiency.} Avg. is computed over representative shared benchmarks in Table~\ref{tab:search-qa-main}: MMSearch, BrowseComp-VL, FVQA, and LiveVQA.}
    \vspace{-3pt}
    \label{tab:training-cost-comparison}
    \scriptsize
    \renewcommand{\arraystretch}{1.18}
    \setlength{\tabcolsep}{3.5pt}
    \resizebox{0.9\textwidth}{!}{%
    \begin{tabular}{l l c c c c c}
        \toprule
        \textbf{Method} & \textbf{Model} & \textbf{SFT Data} & \textbf{RL Data} & \textbf{SFT Cost} & \textbf{RL Cost} & \textbf{Avg.} \\
        \midrule
        OpenSearch-VL~\citep{chen2026opensearch} & 8B dense & 36K & 8K & 256 H20 $\times$ 48h & 64 H20 $\times$ 240h & 58.3 \\
        OpenSearch-VL~\citep{chen2026opensearch} & 30B-A3B & 36K & 8K & 256 H20 $\times$ 96h & 64 H20 $\times$ 240h & 62.6 \\
        \midrule
        \rowcolor[HTML]{F3F5F8}
        \textbf{SimpleSearch-VL} & 8B dense & 5K & 2K & 8 H200 $\times$ 2 h & 8 H200 $\times$ 16 h & 70.3 \\
        \rowcolor[HTML]{F3F5F8}
        \textbf{SimpleSearch-VL} & 30B-A3B & 5K & 2K & 8 H200 $\times$ 3 h & 8 H200  $\times$ 24 h & 74.9 \\
        \bottomrule
    \end{tabular}
    }
\end{table}


\section{Discussion}

Beyond the main SimpleSearch-VL recipe, our codebase also includes two exploratory training options that were implemented but not used in the final reported system. Although they are not part of the main experimental results, they are useful for understanding how rollout selection and interaction efficiency affect multimodal agentic RL.

\paragraph{Mastered-sample filtering.}
The optional \texttt{mastered\_sample\_filter} mechanism filters out prompt groups whose recent rollouts are all correct. This is related in spirit to GRESO-style~\citep{zheng2026act} selective rollouts, which move filtering before generation: if a prompt has recently been consistently all-wrong or all-correct, it is likely to remain zero-variance, so generating a full rollout group only to discard it wastes real computation. The difference is that our filter is simpler and more asymmetric. Rather than deciding whether every prompt should be sampled in the next step, it directly removes mastered all-correct prompts so that the remaining rollout budget is spent on harder examples. This design matches our goal of mining difficult samples. In practice, after easy groups were removed, the average number of interaction rounds increased quickly, suggesting that the policy was exposed to harder prompts that required longer tool-use trajectories.

\paragraph{Round-efficiency reward.}
We also explored a round-efficiency reward that gives an additional bonus to successful trajectories completed with fewer interaction rounds. Concretely, the training loop maintains an online reference for each sample, such as the historical minimum or average number of completion rounds, and rewards a rollout when it solves the sample more efficiently than that reference. This shaping term had the expected behavioral effect: the model learned to shorten successful trajectories and avoid some unnecessary search detours. However, the reduction in rounds did not consistently translate into better final task accuracy. Since multimodal search often benefits from spending extra steps on evidence verification, we leave this reward as an optional engineering setting rather than including it in the main SimpleSearch-VL recipe.

\section{Conclusion}

We introduced SimpleSearch-VL, a simple and efficient framework for multimodal agentic search. The framework combines Factorized Adaptive Rollout for signal-aware RL training, thumbnail-based visual evidence verification for more reliable multimodal retrieval, and goal-conditioned self-summary to keep webpage understanding within the agent itself. Across six multimodal search benchmarks, SimpleSearch-VL substantially improves over untuned Qwen3-VL policies while using only 5K SFT trajectories and 2K RL data, and its 30B-A3B variant remains competitive with agentic Gemini-3-Pro on shared evaluations. These results suggest that carefully designed rollout allocation and verifiable evidence use can be a practical alternative to scaling data, tools, or auxiliary model components for multimodal search agents.

\clearpage
\bibliography{biblio}
\bibliographystyle{plainnat}

\appendix
\clearpage
\section*{Appendix}
\label{sec:appendix}

\makeatletter
\def\addcontentsline#1#2#3{%
  \begingroup
    \let\label\@gobble
    \ifx\@currentHref\@empty
      \phantomsection
    \fi
    \addtocontents{#1}{%
      \protect\contentsline{#2}{#3}{\thepage}{\@currentHref}\protected@file@percent
    }%
  \endgroup
}
\makeatother

\addcontentsline{toc}{part}{Appendix}

\startcontents[appendix]
\begingroup
\hypersetup{linkcolor=deepblue}
\section*{Appendix Contents}
\printcontents[appendix]{}{1}[2]{}
\endgroup

\lstdefinestyle{appendixprompt}{
    basicstyle=\ttfamily\footnotesize,
    breaklines=true,
    columns=fullflexible,
    keepspaces=true,
    showstringspaces=false,
    frame=none
}

\newtcolorbox{appendixbox}[2][]{
    enhanced,
    breakable,
    colback=black!2!white,
    colframe=deepblue!80!black,
    colbacktitle=deepblue!92!black,
    coltitle=white,
    fonttitle=\bfseries,
    boxrule=0.55pt,
    arc=1.2mm,
    left=1.2mm,
    right=1.2mm,
    top=1mm,
    bottom=1mm,
    title=#2,
    #1
}

\section{Implementation Details}
\label{app:implementation-details}

This appendix reports the implementation details of SimpleSearch-VL. We summarize the training data and optimization settings used in our experiments, while omitting framework defaults that are not essential for reproduction.

\subsection{SFT Training Configuration}
\label{app:sft-training-config}

The supervised fine-tuning stage initializes SimpleSearch-VL with executable
evidence-aware trajectories. As shown in
Table~\ref{tab:appendix-sft-config}, both the 8B dense model and the 30B-A3B
MoE model are trained on the same 5,193 trajectories with a 64K-token context
window, three epochs, and batch size 64. The two scales use the corresponding
Qwen3-VL-Instruct checkpoints as initialization and are trained with ms-swift
Megatron SFT on a single H200 node. The 30B-A3B MoE model uses a smaller learning
rate and requires a slightly longer SFT run, but both runs finish within a few
hours, keeping SFT as a lightweight initialization stage rather than the main
source of compute.

\begin{table}[h]
\centering
\small
\setlength{\tabcolsep}{4pt}
\renewcommand{\arraystretch}{1.12}
\caption{\textbf{SFT training configuration.} Recipe-level settings for the 8B dense and 30B-A3B MoE variants; framework defaults are omitted.}
\vspace{-3pt}
\label{tab:appendix-sft-config}
\begin{tabular}{@{}p{0.24\linewidth} >{\centering\arraybackslash}p{0.34\linewidth} >{\centering\arraybackslash}p{0.34\linewidth}@{}}
\toprule
\textbf{Configuration} & \textbf{8B} & \textbf{30B-A3B} \\
\midrule
Base model & \texttt{Qwen3-VL-8B-Instruct} & \texttt{Qwen3-VL-30B-A3B-Instruct} \\
Hardware & 1 H200 node ($8\times$141GB GPUs) & 1 H200 node ($8\times$141GB GPUs) \\
Training time & $\sim$2 hours & $\sim$3 hours \\
Learning rate & $2.0\times10^{-5}$ with minimum LR $5.0\times10^{-7}$ & $1.0\times10^{-5}$ with minimum LR $2.0\times10^{-7}$ \\
\midrule
\rowcolor{gray!6} Training data & \multicolumn{2}{>{\centering\arraybackslash}p{0.68\linewidth}}{5,193 evidence-aware SFT trajectories} \\
\rowcolor{gray!6} Training framework & \multicolumn{2}{>{\centering\arraybackslash}p{0.68\linewidth}}{ms-swift Megatron SFT} \\
\rowcolor{gray!6} Context length & \multicolumn{2}{>{\centering\arraybackslash}p{0.68\linewidth}}{64K tokens} \\
\rowcolor{gray!6} Epochs & \multicolumn{2}{>{\centering\arraybackslash}p{0.68\linewidth}}{3} \\
\rowcolor{gray!6} Batch size & \multicolumn{2}{>{\centering\arraybackslash}p{0.68\linewidth}}{64} \\
\bottomrule
\end{tabular}
\vspace{5pt}
\end{table}

\subsection{RL Training Configuration}
\label{app:rl-training-config}

The reinforcement learning stage further optimizes the SFT-initialized policy in
the same agentic search environment used at inference time. As summarized in
Table~\ref{tab:appendix-rl-config}, RL uses a 64K-token context window for both
model scales and runs for 150 training steps on one H200 node. The training
recipe applies Factorized Adaptive Rollout with \(2\times\) Prompt Expansion
and Rollout Allocation from four to six rollouts, uses a small
\(1.0\times10^{-6}\) learning rate with cosine decay, and disables KL loss. A
large Qwen3-VL judge supplies the answer-level correctness signal used by the
reward and rollout filtering. Compared with SFT, RL is the longer stage
(\(\sim\)16 hours for 8B and \(\sim\)24 hours for 30B-A3B), but it still fits
within a single-node training budget.

\begin{table}[t]
\centering
\small
\setlength{\tabcolsep}{4pt}
\renewcommand{\arraystretch}{1.1}
\caption{\textbf{RL training configuration.} Recipe-level settings for reproducing the 8B dense and 30B-A3B MoE runs.}
\vspace{-3pt}
\label{tab:appendix-rl-config}
\begin{tabular}{@{}p{0.24\linewidth} >{\centering\arraybackslash}p{0.34\linewidth} >{\centering\arraybackslash}p{0.34\linewidth}@{}}
\toprule
\textbf{Configuration} & \textbf{8B} & \textbf{30B-A3B} \\
\midrule
Hardware & 1 H200 node ($8\times$141GB GPUs) & 1 H200 node ($8\times$141GB GPUs) \\
Training time & $\sim$16 hours & $\sim$24 hours \\
\midrule
\rowcolor{gray!6} Context length & \multicolumn{2}{>{\centering\arraybackslash}p{0.68\linewidth}}{64K tokens} \\
\rowcolor{gray!6} Training steps & \multicolumn{2}{>{\centering\arraybackslash}p{0.68\linewidth}}{150} \\
\rowcolor{gray!6} FAR setting & \multicolumn{2}{>{\centering\arraybackslash}p{0.68\linewidth}}{Prompt Expansion with a $2\times$ and Rollout Allocation from 4 to 6 rollouts} \\
\rowcolor{gray!6} Learning rate & \multicolumn{2}{>{\centering\arraybackslash}p{0.68\linewidth}}{$1.0\times10^{-6}$ with cosine decay, 0.05 warmup ratio, and minimum LR $1.0\times10^{-7}$} \\
\rowcolor{gray!6} KL loss & \multicolumn{2}{>{\centering\arraybackslash}p{0.68\linewidth}}{Disabled} \\
\rowcolor{gray!6} Judge model & \multicolumn{2}{>{\centering\arraybackslash}p{0.68\linewidth}}{\texttt{Qwen3-VL-235B-A22B-Instruct}} \\
\bottomrule
\end{tabular}
\vspace{5pt}
\end{table}

\subsection{Training Data Composition}
\label{app:training-data-composition}

\begin{figure}[t]
    \centering
    \includegraphics[width=0.98\linewidth]{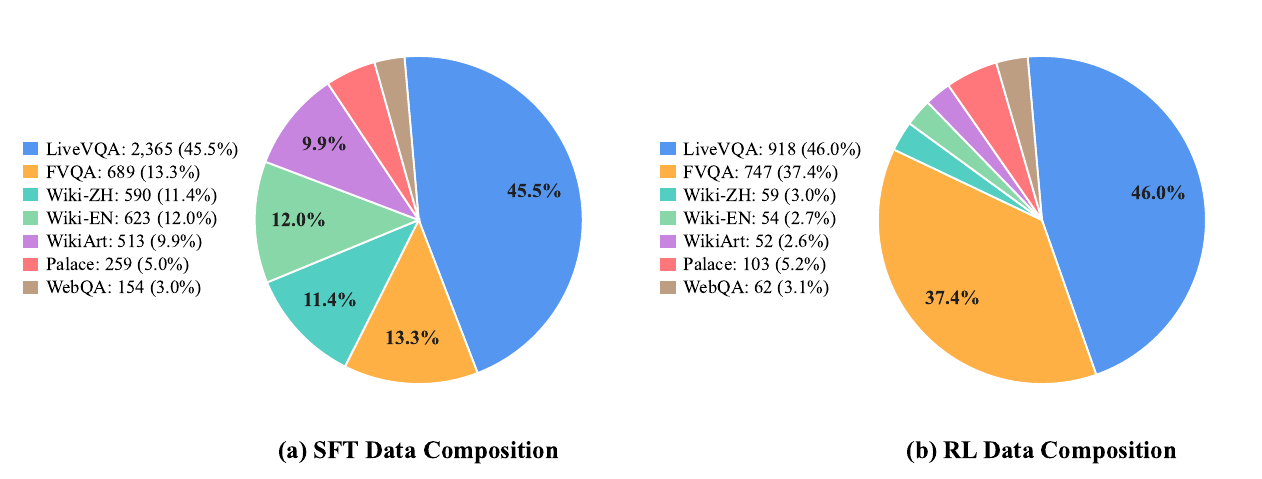}
    \caption{\textbf{Raw source composition of SFT and RL training data.} (a) SFT trajectory mixture. (b) RL data mixture.}
    \label{fig:appendix-data-source-pies}
    \vspace{10pt}
\end{figure}

Fig.~\ref{fig:appendix-data-source-pies} summarizes the raw data-source
composition used by the SFT and RL stages. The final SFT set contains 5,193
evidence-aware trajectories, led by LiveVQA (2,365) and complemented by FVQA
(689), Wiki-EN (623), Wiki-ZH (590), WikiArt (513), Palace (259), and WebQA
(154). The RL pool contains 1,995 prompts and is intentionally concentrated on
the sources that provide reliable answer checking and useful search difficulty:
LiveVQA contributes 918 prompts and FVQA contributes 747, while Palace, WebQA,
Wiki-ZH, Wiki-EN, and WikiArt provide smaller but diverse long-tail coverage.
This mixture gives SFT broad evidence-aware behavior and reserves RL for a
cleaner prompt pool with stable binary feedback.

\begin{figure}[t]
    \centering
    \includegraphics[width=0.98\linewidth]{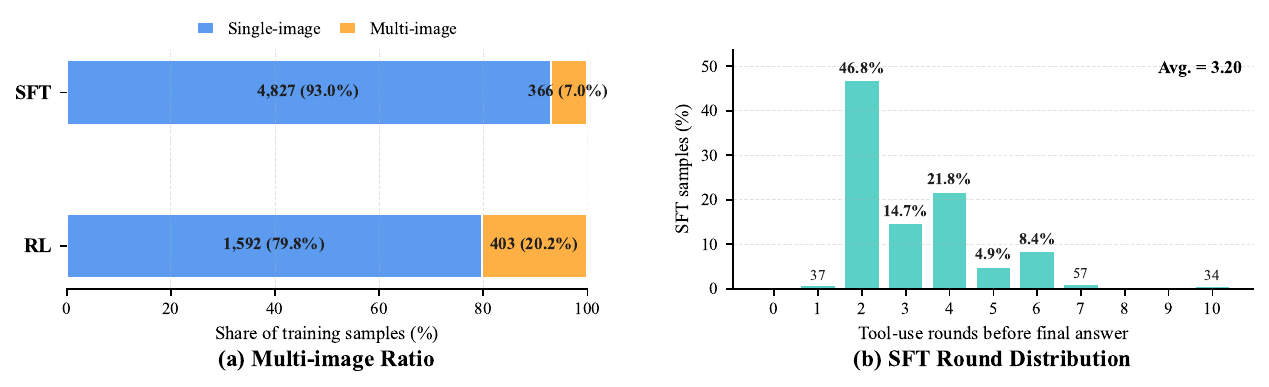}
    \caption{\textbf{Input-image type and SFT round distribution.} (a) Single-image and multi-image shares in SFT and RL data. (b) SFT tool-use round distribution, where a round denotes one tool call before the final answer.}
    \label{fig:appendix-data-diagnostics}
    \vspace{10pt}
\end{figure}

Fig.~\ref{fig:appendix-data-diagnostics} provides two complementary views of the training data. \textbf{Multi-image coverage.} We count multi-image samples by the number of original input images, excluding retrieved visual thumbnails used for evidence verification. Under this definition, the SFT archive contains 366 multi-image trajectories out of 5,193 samples (7.0\%), and the RL pool contains 403 multi-image prompts out of 1,995 samples (20.2\%). All multi-image samples are from LiveVQA, which supplies the main supervision signal for multi-image search and comparison. \textbf{Trajectory length.} The SFT trajectories use 3.20 tool-use rounds on average before the final answer, with most samples concentrated between two and four rounds. This distribution indicates that the SFT data emphasizes concise tool-interleaved reasoning rather than unnecessarily long search chains.

\section{Agent Workflow and Tool Interface}
\label{app:agent-workflow}

The agent operates in a tool-interleaved loop. Given a question and one or more input images, the system assigns explicit image identifiers, e.g., \texttt{img\_idx=0} and \texttt{img\_idx=1}. At every turn, the model emits a \texttt{<thinking>} block and then takes exactly one action: a JSON-formatted tool call or a final \texttt{<answer>}. Tool observations are appended to the context, and the loop continues until a valid answer is produced or the configured budget is reached.

\begin{table}[H]
\centering
\vspace{5pt}
\caption{\textbf{Tool declarations in SimpleSearch-VL.} The table lists each tool's role, input signature, and returned evidence fields.}
\vspace{-3pt}
\renewcommand{\arraystretch}{1.18}
\resizebox{\linewidth}{!}{%
\begin{tabular}{c
>{\centering\arraybackslash}m{0.22\linewidth}
>{\centering\arraybackslash}m{0.27\linewidth}
>{\centering\arraybackslash}m{0.36\linewidth}}
\toprule
\textbf{Tool} & \textbf{Role} & \textbf{Input Signature} & \textbf{Output} \\
\midrule
\texttt{text\_search} &
Search the web for external textual knowledge. &
\texttt{query}: list of search queries &
\texttt{title}: webpage title\newline \texttt{url}: source webpage\newline \texttt{snippet}: short matched context \\
\midrule
\texttt{image\_search} &
Perform region-level reverse image search using a selected image region as the visual query. &
\texttt{regions}: list of \{\texttt{img\_idx}, \texttt{bbox\_2d}\} &
\texttt{thumbnail}: visual match preview\newline \texttt{title}: matched webpage title\newline \texttt{url}: source webpage \\
\midrule
\texttt{visit} &
Extract webpages relevant to a goal. &
\texttt{url}: list of webpages\newline \texttt{goal}: extraction goal string &
\texttt{summary}: goal-conditioned summary \\
\bottomrule
\end{tabular}}
\label{tab:search-tools}
\end{table}

\paragraph{Text Search.}
The text-search tool is used when the model already has keywords or candidate entities to verify. Its input is a list of search queries, and its output contains webpage titles, URLs, and short matched snippets. The returned snippets provide preliminary evidence; when exact webpage evidence is needed, the model is instructed to call \texttt{visit} on the selected URLs.

\paragraph{Image Search.}
The image-search tool performs region-level reverse image search, using selected image regions as visual queries. Its input is a list of regions, where each region contains an \texttt{img\_idx} and a normalized \([0,1000]\) \texttt{bbox\_2d}; \([0,0,1000,1000]\) denotes the full image. Its output includes the matched thumbnail, webpage title, and source URL. Unlike single-image search interfaces, our schema allows a single batched call to include regions from multiple input images, which is important for multi-image visual questions. The returned thumbnails are used for visual evidence verification before the agent trusts the associated title or URL.

\paragraph{Visit.}
The visit tool takes a list of URLs and a goal string as input. For each URL, a webpage reader first fetches and normalizes the content into text; the agent then produces a concise summary conditioned on the goal. This keeps webpage understanding inside the model's own generation process while avoiding an additional deployed summary model during training and inference. The exact self-summary prompt is provided in Table~\ref{tab:appendix-self-summary-prompt}.

\section{Evidence-Aware Reasoning Data Construction}
\label{app:evidence-aware-data}

We construct SFT trajectories with an evidence-aware data pipeline. The pipeline is designed to produce trajectories where visual observations, image-search thumbnails, web evidence, and final answers are explicitly connected. We focus on the three model-involved stages after candidate pool construction: direct-answer filtering, agentic rollout filtering, and trajectory audit with reasoning rewrite.

\paragraph{Step 0: Candidate Pool Construction.}
We first collect candidate visual QA samples from FVQA, LiveVQA, Wiki-EN, Wiki-ZH, WikiArt, Palace, and WebQA. Since RL training requires a reliable correctness signal, we focus on questions with a \emph{determinate and verifiable answer boundary}: the expected answer should be a specific entity, attribute, event, count, date, title, location, relation, or another objective fact. Open-ended or non-factoid queries, such as questions asking for broad descriptions, explanations, opinions, or ambiguous ``why/how'' reasoning, are removed because many plausible responses could be acceptable and cannot be judged robustly by a binary reward.

\paragraph{Step 1: Direct-Answer Filtering.}
We use \texttt{Qwen3-VL-235B-A22B-Instruct} as both the direct-answer model and the judge. The model first produces a direct answer for each sample. The same model then performs two judgments: whether the prediction matches the reference answer, and whether the question is a clear fact-seeking query with an objective correctness criterion. Samples that can be answered correctly in this direct-answer setting are removed, and samples judged ambiguous, subjective, open-ended, or insufficiently contextualized are also excluded. Only incorrect, valid, fact-seeking samples are kept for agentic rollout. This stage filters out directly answerable examples while preserving questions that require external evidence.

\paragraph{Step 2: Agentic Rollout Filtering.}
We then generate tool-interleaved trajectories for the retained samples using \texttt{Qwen3-VL-235B-A22B-Instruct} as the policy model. The rollout environment provides three tools, \texttt{text\_search}, \texttt{image\_search}, and \texttt{visit}; image-search observations include visual thumbnails for evidence verification. Sampling is deterministic during data construction (\(T=0\)), and each sample is allowed up to 10 model turns. We keep trajectories that end with a valid \texttt{<answer>}, follow the required assistant format, are judged answer-correct, and contain substantive tool use. Trajectories that fail format checks, terminate without a supported final answer, or do not use external evidence are rejected.

\paragraph{Step 3: Joint Trajectory Audit and Reasoning Rewrite.}
We use \texttt{gemini-3.1-pro} to perform trajectory audit and reasoning rewrite in a single model call. The auditor rejects trajectories with unsupported final answers, repeated low-value searches, premature or unnecessary tool calls, or weak connections between the visual evidence and the final response. For retained samples, Gemini rewrites only the assistant reasoning and final answer while preserving the original tool calls, tool responses, and evidence order. The rewrite explicitly requires image-search reasoning to compare retrieved thumbnails with the queried image region before trusting the associated title or URL, making visual evidence verification visible in the SFT trajectory.

\section{Prompt Templates}
\label{app:prompt-templates}

This section groups the prompt templates used by SimpleSearch-VL into the agent system prompt, tool prompt, self-summary prompt, and judge prompt. The same agent and tool prompts are used during training and inference.

\subsection{Agent System Prompt}
\label{app:system-prompt}

\begin{center}
\captionof{table}{\textbf{Agent system prompt.} The active date is replaced by \texttt{\{current\_date\}}.}
\label{tab:appendix-agent-prompt}
\vspace{1mm}
\begin{appendixbox}{Agent System Prompt}
\begin{lstlisting}[style=appendixprompt]
#Role
You are a visual research assistant.
Given a question, your task is to solve the problem **one substep at a time**.

## Guiding Principles
At each turn, you must **either**:
1. Issue **one specific tool** enclosed in <tool_call></tool_call> tags,
2. Or provide the **final answer** enclosed in <answer></answer> tags.

All outputs **must begin with a thought** enclosed in <thinking></thinking> tags, explaining your current reasoning and what to do next.

## Output Format (strict):
Always start with <thinking>. Do not output the previous reasoning chain. Then, depending on the case, output one of the following:

1. If reasoning continues:
<thinking>Your current reasoning and next plan</thinking>
<tool_call>One precise tool call to assist your reasoning</tool_call>

2. If ready to conclude:
<thinking>Summarize all reasoning and derive the answer</thinking>
<answer>Final answer</answer>
\end{lstlisting}
\end{appendixbox}
\end{center}

\subsection{Tool Prompt and Schemas}
\label{app:tool-definition-prompts}

The tool section is appended after the system prompt and before the current date. It exposes only the active tool schemas and enforces a single valid JSON tool call per assistant turn.

\begin{center}
\captionof{table}{\textbf{Tool calling prompt.} Function schemas are injected into \texttt{\{tool\_json\_lines\}}.}
\label{tab:appendix-tool-prompt}
\vspace{1mm}
\begin{appendixbox}{Tool Calling Prompt}
\begin{lstlisting}[style=appendixprompt]
# Tools

You may call exactly one function per assistant turn. Use batched arguments inside that single tool call when you need parallel search.

You are provided with function signatures within <tools></tools> XML tags:
<tools>
{tool_json_lines}
</tools>

For the function call, return one valid JSON object inside <tool_call></tool_call>.
The top-level object must have exactly two sibling keys: "name" and "arguments".
Never insert an extra object between "name" and "arguments".

Use these exact JSON shapes:
<tool_call>
{"name": "text_search", "arguments": {"query": ["example query"]}}
</tool_call>
<tool_call>
{"name": "visit", "arguments": {"url": ["https://example.com"], "goal": "find relevant evidence"}}
</tool_call>
<tool_call>
{"name": "image_search", "arguments": {"regions": [{"img_idx": 0, "bbox_2d": [0, 0, 1000, 1000]}]}}
</tool_call>

Current date: {current_date}
\end{lstlisting}
\end{appendixbox}
\end{center}

The active schemas are listed below for readability.

\begin{center}
\captionof{table}{\textbf{\texttt{image\_search} schema.} Multi-image region search is supported through \texttt{img\_idx}.}
\label{tab:appendix-image-search-schema}
\vspace{1mm}
\begin{appendixbox}{Tool Definition: image\_search}
\begin{lstlisting}[style=appendixprompt]
Description:
Perform reverse image search for one or more input image regions. Use this to identify unfamiliar objects, logos, landmarks, artworks, products, people, or scenes. Input images are referred to by img_idx, starting from 0. Each region must pair one img_idx with one bbox_2d in 0-1000 scale. Use [0, 0, 1000, 1000] for a whole image. The tool response labels each searched region with the same img_idx.

Arguments:
{
  "regions": [
    {
      "img_idx": number,
      "bbox_2d": [x1, y1, x2, y2]
    }
  ]
}

Constraints:
- 1 to 3 regions per call.
- bbox_2d uses the 0-1000 coordinate system.
- [0, 0, 1000, 1000] denotes the full image.

Example:
<tool_call>
{
  "name": "image_search",
  "arguments": {
    "regions": [
      {"img_idx": 0, "bbox_2d": [0, 0, 1000, 1000]}
    ]
  }
}
</tool_call>
\end{lstlisting}
\end{appendixbox}
\end{center}

\begin{center}
\captionof{table}{\textbf{\texttt{text\_search} schema.} Complementary queries can be batched in one call.}
\label{tab:appendix-text-search-schema}
\vspace{1mm}
\begin{appendixbox}{Tool Definition: text\_search}
\begin{lstlisting}[style=appendixprompt]
Description:
Perform text-based web searches and return top results. Use when you already have specific keywords or facts to look up. Supply complementary queries in a single call when useful.

Arguments:
{
  "query": ["search query 1", "search query 2", "..."]
}

Constraints:
- 1 to 3 queries per call.
- The query array must stay inside arguments.

Example:
<tool_call>
{
  "name": "text_search",
  "arguments": {
    "query": ["example query"]
  }
}
</tool_call>
\end{lstlisting}
\end{appendixbox}
\end{center}

\begin{center}
\captionof{table}{\textbf{\texttt{visit} schema.} Webpages are summarized with respect to an extraction goal.}
\label{tab:appendix-visit-schema}
\vspace{1mm}
\begin{appendixbox}{Tool Definition: visit}
\begin{lstlisting}[style=appendixprompt]
Description:
Visit webpages and extract evidence relevant to your goal. Returns extracted evidence, a summary, and related links for further research. Use exact, complete URLs from search results or related links. Do not modify, truncate, or guess URLs.

Arguments:
{
  "url": ["https://example.com"],
  "goal": "what information to extract from the webpages"
}

Constraints:
- 1 to 3 URLs per call by default; the maximum can be configured at runtime.
- URLs must be exact and complete.
- goal should specify the evidence needed for answering the question.

Example:
<tool_call>
{
  "name": "visit",
  "arguments": {
    "url": ["https://example.com"],
    "goal": "find relevant evidence"
  }
}
</tool_call>
\end{lstlisting}
\end{appendixbox}
\end{center}

\begin{center}
\captionof{table}{\textbf{Image-search observation instruction.} Visual thumbnails are checked before using titles or URLs as evidence.}
\label{tab:appendix-thumbnail-instruction}
\vspace{1mm}
\begin{appendixbox}{Visual Thumbnail Verification Instruction}
\begin{lstlisting}[style=appendixprompt]
Reminder: Each image thumbnail corresponds to the result immediately below it.
Compare the thumbnails with the queried image region first.
Visit all URLs whose thumbnails and titles are visually and semantically consistent with the target.
If thumbnails do not match the target region, treat those results as low-confidence and do not rely on them.
You can pass multiple URLs in one visit call.
\end{lstlisting}
\end{appendixbox}
\end{center}

\begin{center}
\captionof{table}{\textbf{Multi-image image-search example.} One call can search localized regions from different input images.}
\label{tab:appendix-multi-image-prompt}
\vspace{1mm}
\begin{appendixbox}{Multi-Image Search Example}
\begin{lstlisting}[style=appendixprompt]
Input images are indexed from 0. A single image_search call may include regions from multiple input images:

<tool_call>
{
  "name": "image_search",
  "arguments": {
    "regions": [
      {"img_idx": 0, "bbox_2d": [0, 0, 1000, 1000]},
      {"img_idx": 1, "bbox_2d": [120, 180, 760, 920]}
    ]
  }
}
</tool_call>

The returned observation keeps the same img_idx for each searched region. If thumbnails are returned, compare each thumbnail against the queried region before trusting the corresponding title or URL.
\end{lstlisting}
\end{appendixbox}
\end{center}

\subsection{Self-Summary Prompt}
\label{app:self-summary-prompt}

After \texttt{visit} fetches and normalizes a webpage, the policy model itself summarizes the content with respect to the extraction goal.

\begin{center}
\captionof{table}{\textbf{Self-summary prompt.} The model summarizes webpage content with respect to the extraction goal.}
\label{tab:appendix-self-summary-prompt}
\vspace{1mm}
\begin{appendixbox}{Self-Summary Prompt for visit}
\begin{lstlisting}[style=appendixprompt]
System:
You are a helpful assistant. Your task is to summarize webpage content concisely and factually.

User:
Summarize the following webpage content with respect to the user's goal.

Webpage Content (may be truncated):
{webpage_content}

User Goal: {goal}

Instructions:
- Summarize the main content of the webpage in no more than five sentences.
- Your summary should cover the overall key points of the page, not just parts related to the user's goal.
- If any part of the content is helpful for answering the user's goal, include it clearly in the summary.
- If the page contains no information relevant to the goal, state that briefly.
- Your summary should be concise, factual, and informative.
- Return the summary text directly. Do not use JSON, code fences, or special formatting.
\end{lstlisting}
\end{appendixbox}
\end{center}

\subsection{Judge Prompt}
\label{app:judge-prompt}

The final training configuration uses a binary accuracy reward. Exact matching is attempted first; otherwise, the fallback LLM judge uses the prompt below.

\begin{center}
\captionof{table}{\textbf{Accuracy judge prompt.} The fallback judge returns the binary correctness signal used by the reward.}
\label{tab:appendix-judge-prompt}
\vspace{1mm}
\begin{appendixbox}{Accuracy Judge Prompt}
\begin{lstlisting}[style=appendixprompt]
You are an impartial judge evaluating whether a deep research report contains the correct answer.

[Question]
{question}

[Correct Answer]
{reference_answer}

[Deep Research Report]
{assistant_answer}

Task: Determine if the deep research report contains the correct answer anywhere in its content.

Instructions:
1. Read through the entire research report carefully.
2. Look for the correct answer anywhere in the report. It may be embedded in paragraphs, tables, or sections.
3. Check if the information in the report is consistent with the correct answer.
4. The answer does NOT need to be in a specific format or labeled as "final answer".
5. Provide your reasoning.
6. Answer with "yes" if the report contains the correct answer, and "no" if it does not or contradicts it.

Output format:
correct: [yes/no]
reasoning: [your explanation]
\end{lstlisting}
\end{appendixbox}
\end{center}

\section{Visualization}
\label{app:visualization}

\subsection{Standard Single-Image Cases}
\label{app:vis-standard}

\begin{figure}[p]
    \centering
    \includegraphics[width=\linewidth,height=0.88\textheight,keepaspectratio]{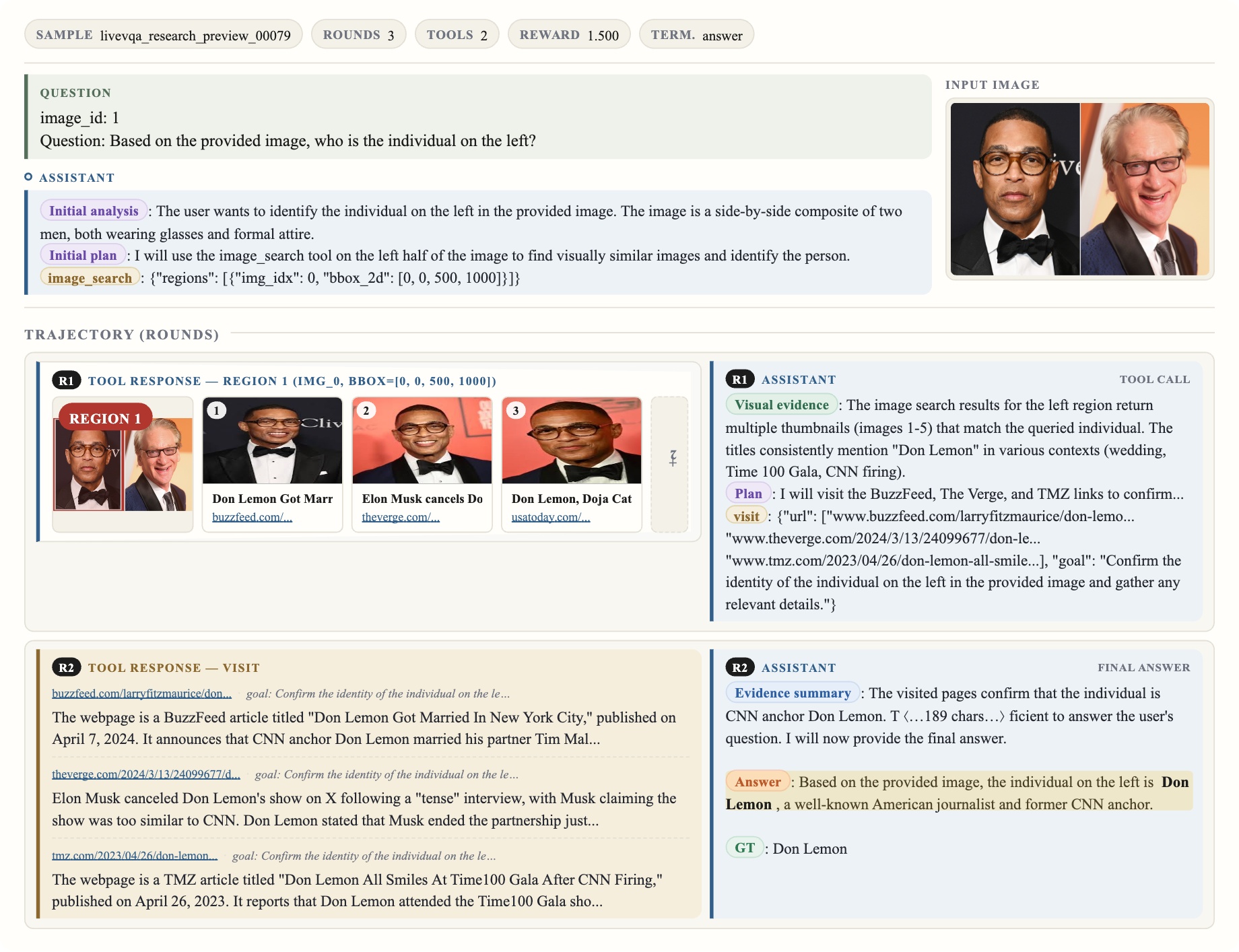}
    \caption{\textbf{Standard single-image case.} The agent searches the left image region, matches retrieved thumbnails and titles to Don Lemon, and verifies the identity with visited webpages before answering.}
    \label{fig:appendix-vis-standard-livevqa}
    \vspace{10pt}
\end{figure}

\begin{figure}[p]
    \centering
    \includegraphics[width=\linewidth,height=0.88\textheight,keepaspectratio]{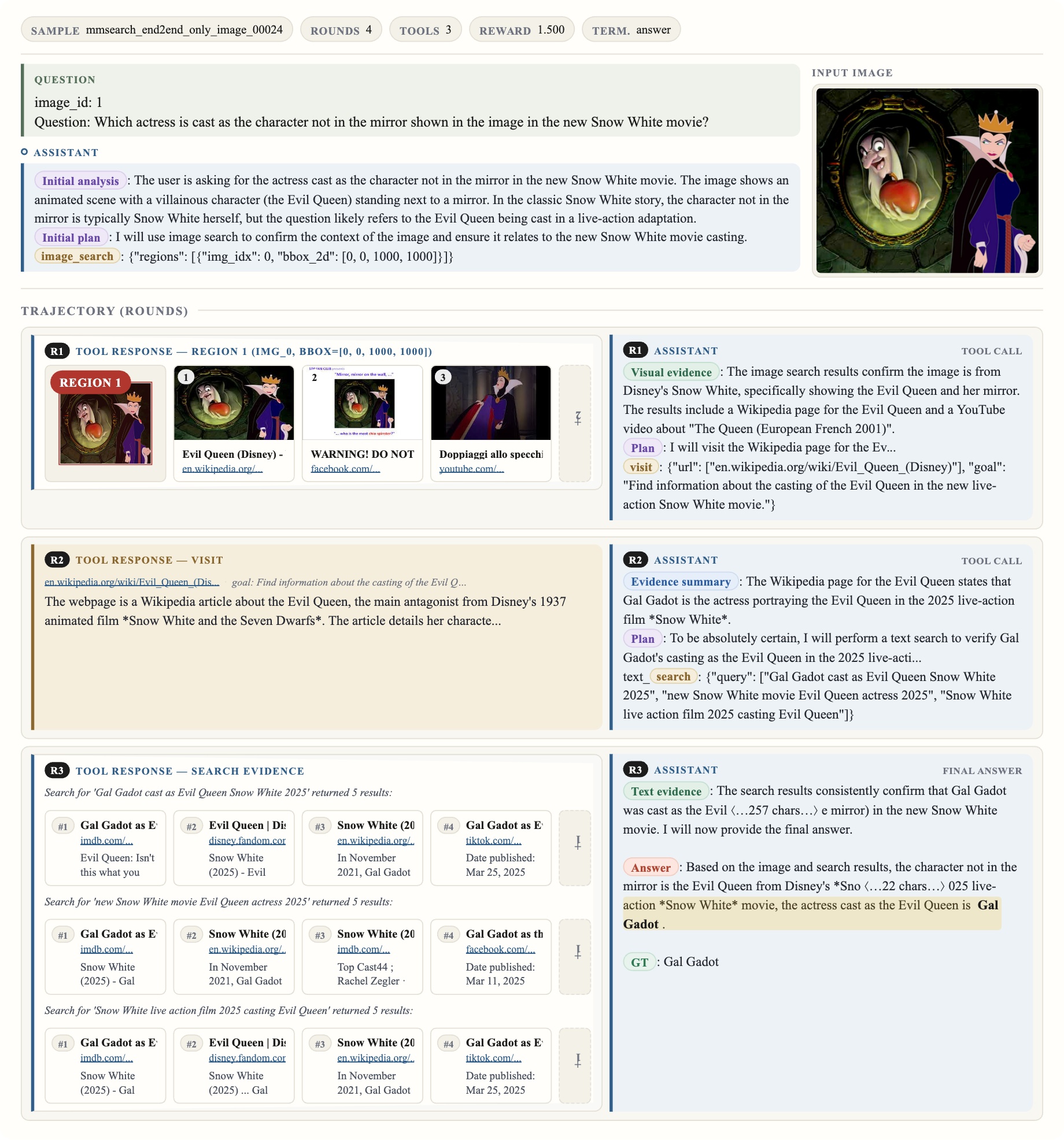}
    \caption{\textbf{Standard single-image case with visual-to-text verification.} The reverse-image-search result grounds the scene as the Evil Queen from \emph{Snow White}; webpage and text evidence then confirm Gal Gadot as the actress in the new film.}
    \label{fig:appendix-vis-standard-snowwhite}
    \vspace{10pt}
\end{figure}

Fig.~\ref{fig:appendix-vis-standard-livevqa} and Fig.~\ref{fig:appendix-vis-standard-snowwhite} show standard single-image trajectories. In Fig.~\ref{fig:appendix-vis-standard-livevqa}, the agent crops the left half of a composite news image, retrieves visually matching thumbnails for Don Lemon, and then visits supporting pages before answering. In Fig.~\ref{fig:appendix-vis-standard-snowwhite}, reverse image search first grounds the animated scene as Disney's Evil Queen and magic mirror, after which webpage evidence and text search verify that Gal Gadot plays the Evil Queen in the 2025 live-action \emph{Snow White}. These examples show the basic evidence chain used by SimpleSearch-VL: region-level reverse image search proposes an entity, thumbnail consistency filters the proposal, and text or webpage evidence confirms the final answer.

\subsection{Single-Image Case without Image Search}
\label{app:vis-no-image-search}

\begin{figure}[p]
    \centering
    \includegraphics[width=\linewidth,height=0.88\textheight,keepaspectratio]{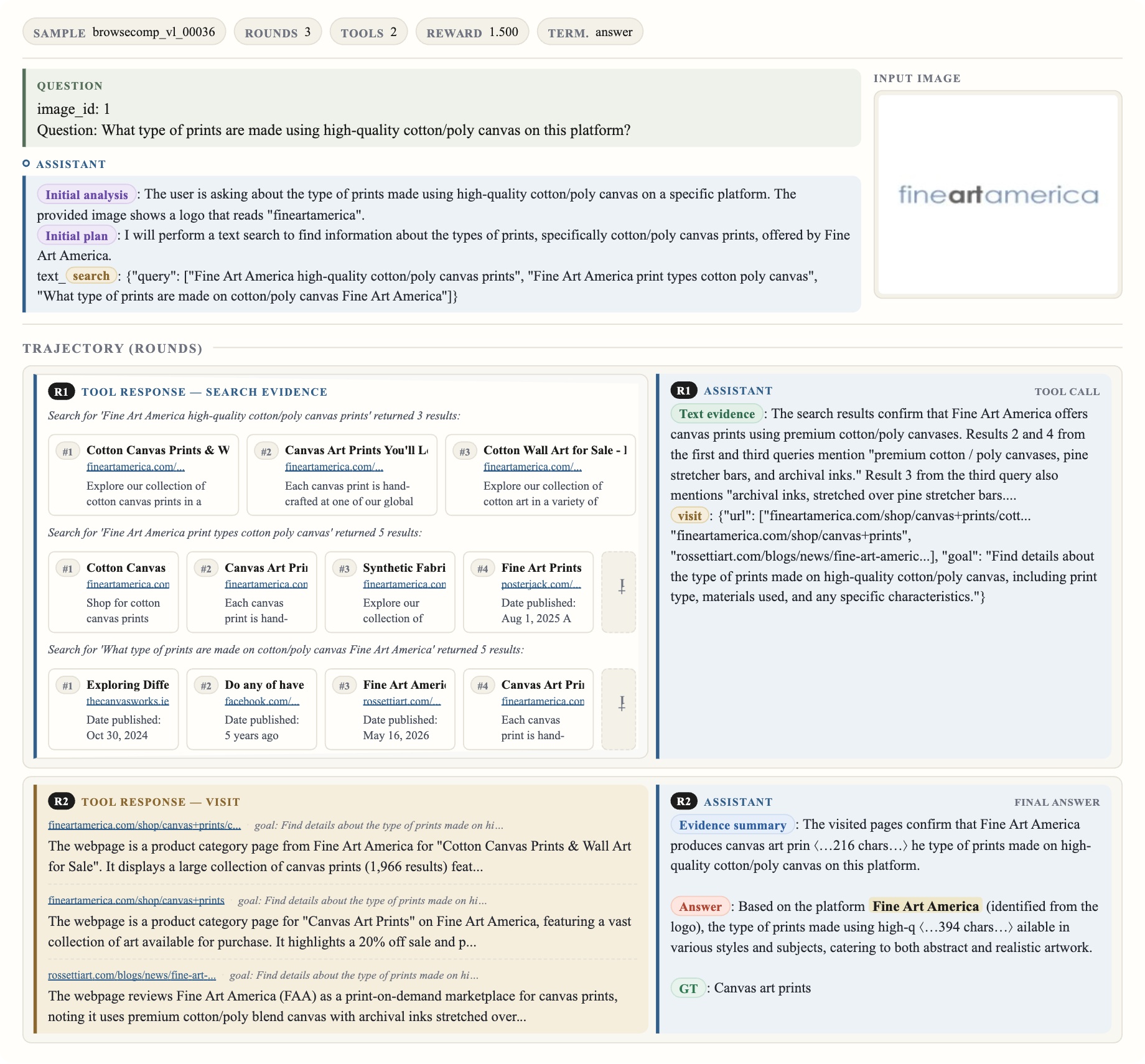}
    \caption{\textbf{Single-image case without image search.} For this straightforward logo-based question, the model does not call image search; it uses the visible Fine Art America text as an anchor and verifies the answer through text search and webpage visiting.}
    \label{fig:appendix-vis-nf-fineartamerica}
    \vspace{10pt}
\end{figure}

Fig.~\ref{fig:appendix-vis-nf-fineartamerica} shows a simple case where the model chooses not to use image search. The input image contains a readable platform logo, ``fineartamerica'', and the question asks about the type of prints made from high-quality cotton/poly canvas on that platform. Since this visual cue is already sufficient to identify the platform, the agent skips reverse image search, directly issues text-search queries about Fine Art America canvas prints, visits the relevant product pages, and verifies that the platform sells canvas art prints made with premium cotton/poly blend canvas, archival inks, and stretcher bars. This example illustrates adaptive tool allocation: when the image already supplies a reliable textual anchor and the problem is straightforward, the model can avoid unnecessary reverse image search and move directly to textual evidence gathering.

\subsection{Multi-Frame Cases}
\label{app:vis-multi-frame}

\begin{figure}[p]
    \centering
    \includegraphics[width=\linewidth,height=0.88\textheight,keepaspectratio]{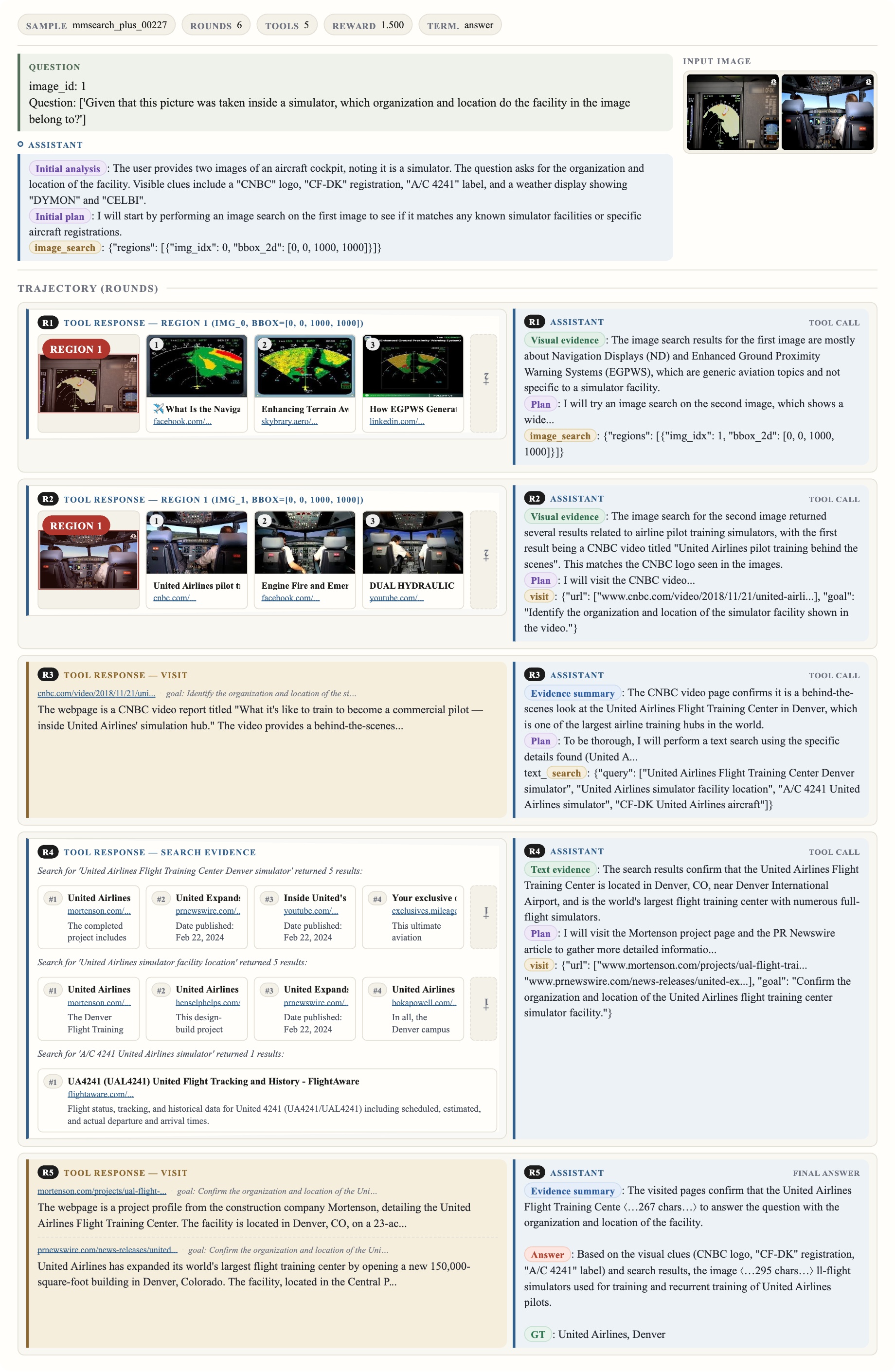}
    \caption{\textbf{Multi-frame case with frame selection.} The first cockpit frame gives generic simulator evidence, while the second frame retrieves a United Airlines training video; the agent then verifies that the facility is the United Airlines Flight Training Center in Denver.}
    \label{fig:appendix-vis-mf-simulator}
    \vspace{10pt}
\end{figure}

\begin{figure}[p]
    \centering
    \includegraphics[width=\linewidth,height=0.88\textheight,keepaspectratio]{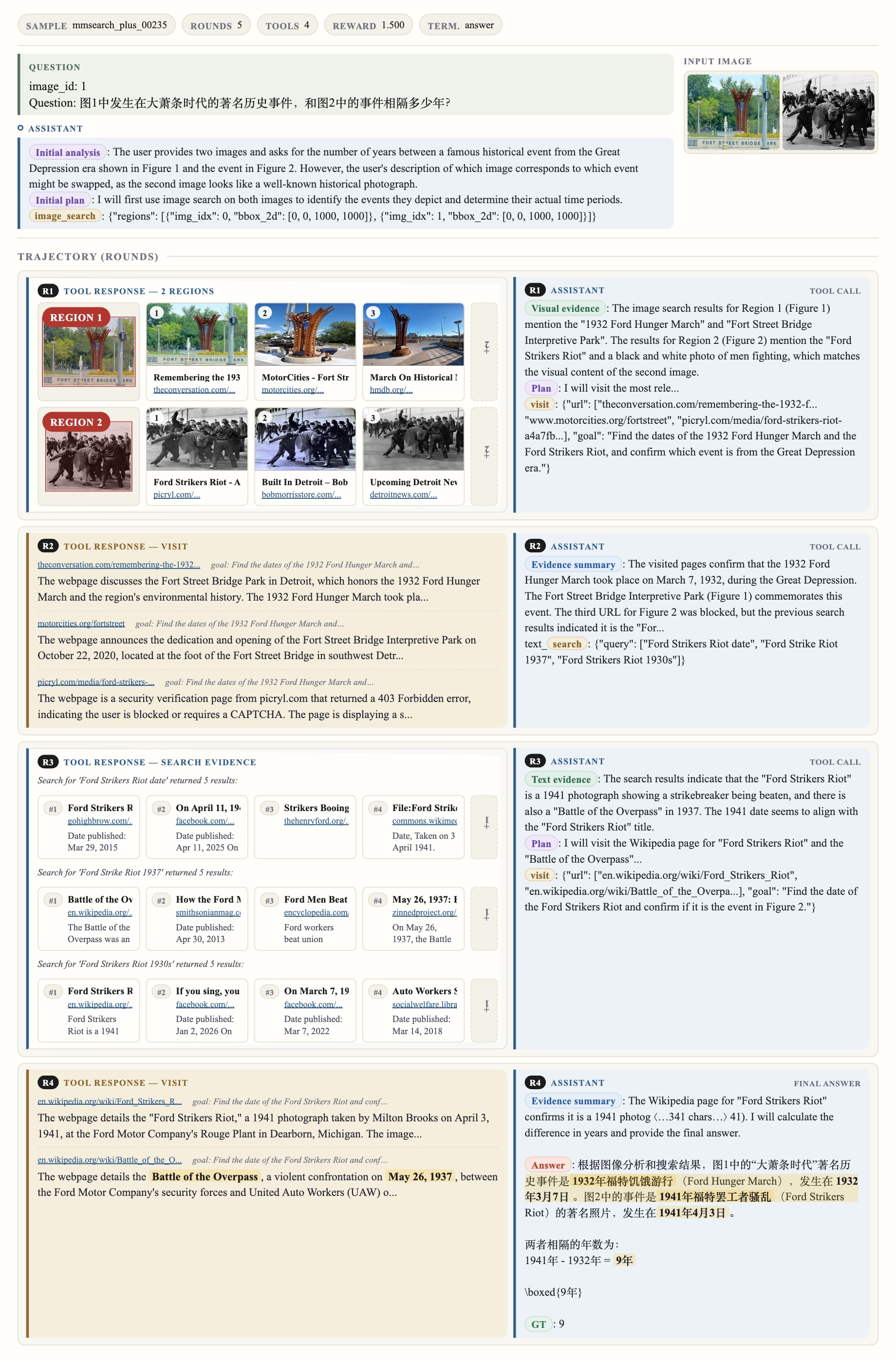}
    \caption{\textbf{Multi-frame case with cross-image comparison.} The agent links the first frame to the 1932 Ford Hunger March and the second frame to the 1941 Ford Strikers Riot, verifies both dates through webpage evidence, and computes the correct nine-year interval.}
    \label{fig:appendix-vis-mf-history}
    \vspace{10pt}
\end{figure}

Fig.~\ref{fig:appendix-vis-mf-simulator} and Fig.~\ref{fig:appendix-vis-mf-history} illustrate multi-frame reasoning. The simulator example first searches one cockpit image and obtains generic aviation evidence, then switches to the second frame, where the CNBC-style visual match identifies a United Airlines pilot-training video. The agent visits and cross-checks this evidence to answer that the facility belongs to United Airlines in Denver. The historical-event example searches both frames in one visual step: the first image is linked to the 1932 Ford Hunger March, while the second is linked to the 1941 Ford Strikers Riot. The final answer is obtained by verifying both dates and computing the nine-year gap. These trajectories show why the interface exposes image indices explicitly: the model can decide which frame is informative, and can compare evidence across frames when the question requires relational reasoning.

\subsection{Multi-Region Case}
\label{app:vis-multi-region}

\begin{figure}[p]
    \centering
    \includegraphics[width=\linewidth,height=0.88\textheight,keepaspectratio]{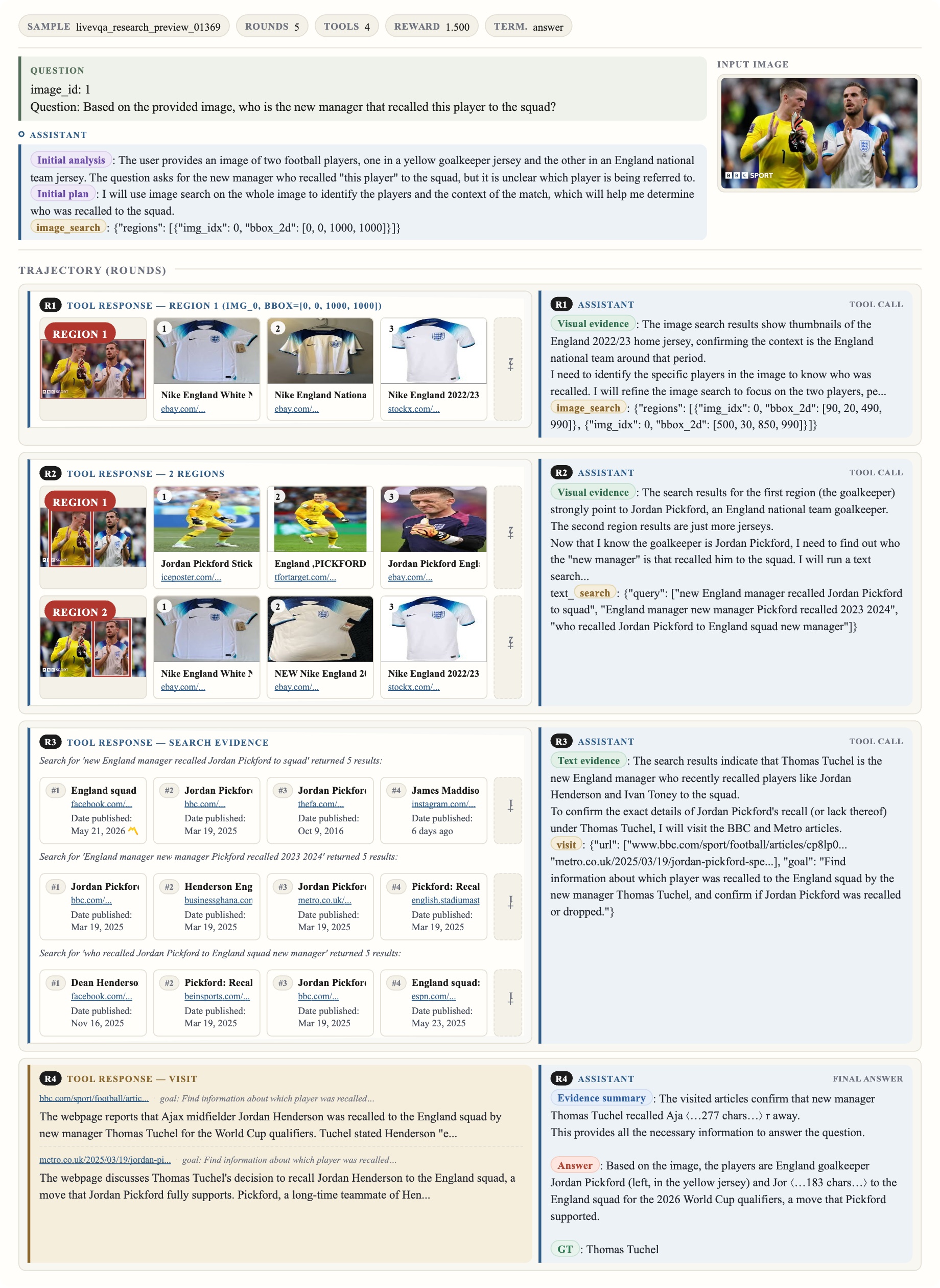}
    \caption{\textbf{Multi-region case.} After a broad whole-image reverse search, the agent searches localized player regions, uses the goalkeeper match to anchor the England-squad context, and verifies from webpages that Thomas Tuchel recalled Jordan Henderson.}
    \label{fig:appendix-vis-mr-football}
    \vspace{10pt}
\end{figure}

Fig.~\ref{fig:appendix-vis-mr-football} shows a trajectory that searches multiple regions within the same image. The full-image query only establishes a broad England-football context, so the agent refines the visual query by cropping the goalkeeper and the second player separately. The first region identifies Jordan Pickford, while the subsequent textual and webpage evidence clarifies that the relevant recall story concerns Jordan Henderson under new England manager Thomas Tuchel. This example demonstrates the role of region-level reverse image search: rather than trusting a whole-image match, the model can isolate entities and then use text evidence to resolve the question-specific relation.

\subsection{Failure Cases}
\label{app:vis-failure-cases}

\begin{figure}[p]
    \centering
    \includegraphics[width=\linewidth,height=0.88\textheight,keepaspectratio]{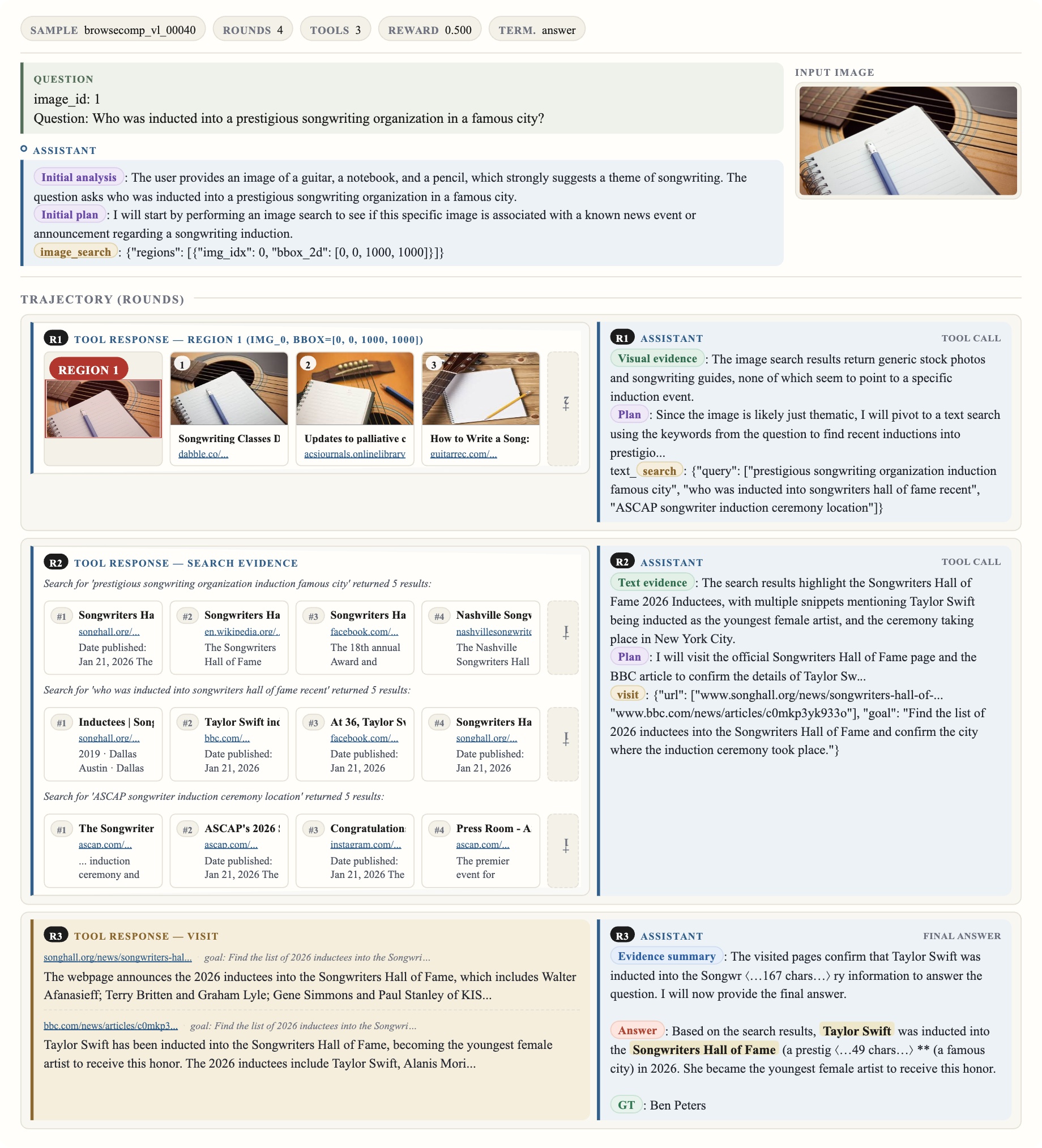}
    \caption{\textbf{Failure case with wrong entity selection.} A generic songwriting image leads the agent to search for a plausible Songwriters Hall of Fame induction and answer Taylor Swift, although the benchmark target is Ben Peters.}
    \label{fig:appendix-vis-failure-songhall}
    \vspace{10pt}
\end{figure}

\begin{figure}[p]
    \centering
    \includegraphics[width=\linewidth,height=0.88\textheight,keepaspectratio]{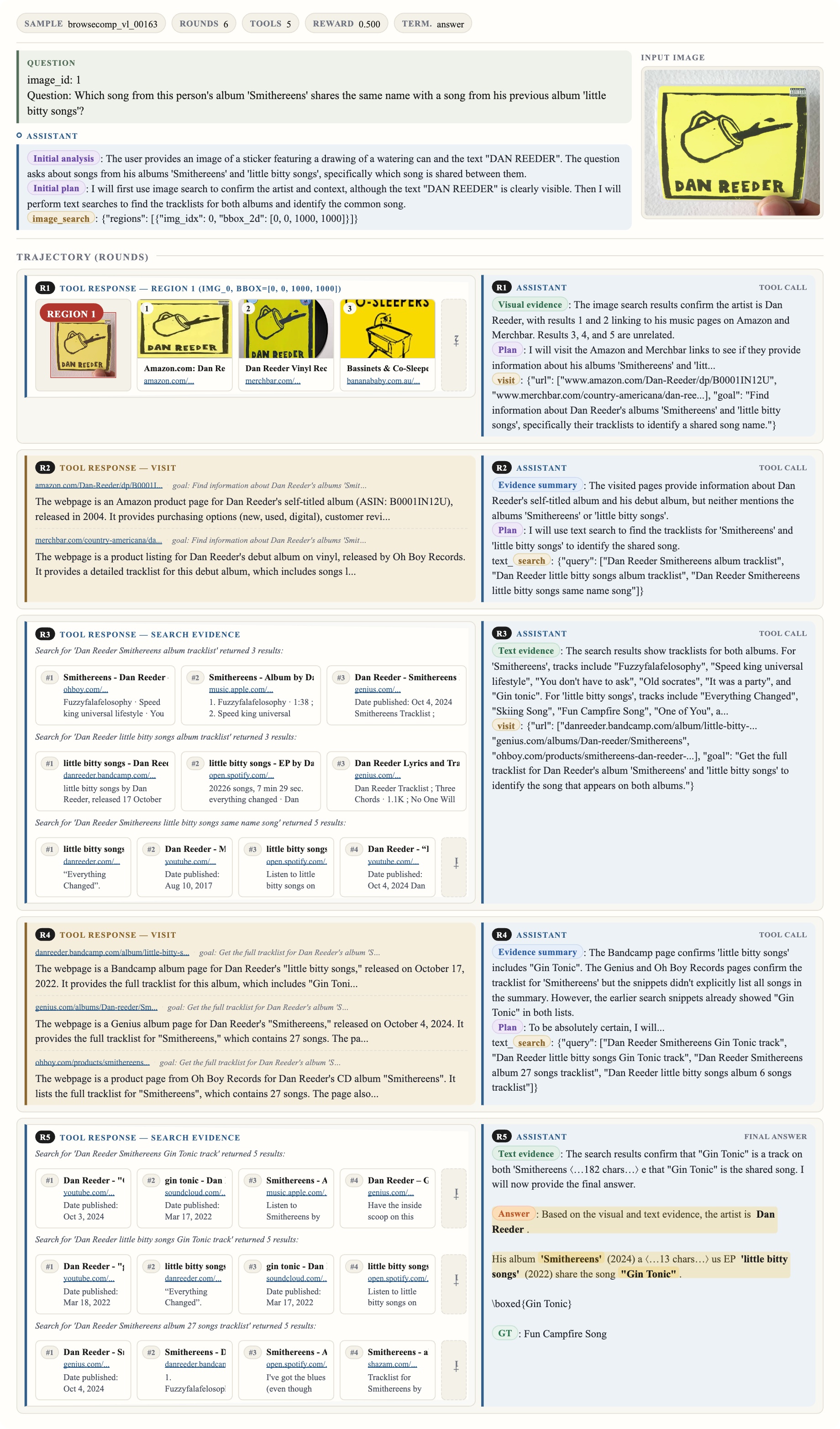}
    \caption{\textbf{Failure case after correct visual grounding.} Reverse image search identifies Dan Reeder, but the later track-list comparison selects ``Gin Tonic'' instead of the target shared song, ``Fun Campfire Song''.}
    \label{fig:appendix-vis-failure-danreeder}
    \vspace{10pt}
\end{figure}

Fig.~\ref{fig:appendix-vis-failure-songhall} and Fig.~\ref{fig:appendix-vis-failure-danreeder} show two failure modes. In Fig.~\ref{fig:appendix-vis-failure-songhall}, the image is only a thematic cue for songwriting, and reverse image search returns generic songwriting pages rather than the intended referent. The agent then follows a plausible but wrong text-search path to Taylor Swift's Songwriters Hall of Fame induction, whereas the benchmark target is Ben Peters. This is an entity-selection failure: the external evidence is internally consistent, but it resolves the wrong hidden target. In Fig.~\ref{fig:appendix-vis-failure-danreeder}, the visible sticker correctly identifies Dan Reeder, and the agent retrieves track-list evidence for \emph{Smithereens} and \emph{little bitty songs}. However, it selects ``Gin Tonic'' as the overlapping song, while the gold answer is ``Fun Campfire Song''. This is a fine-grained evidence extraction failure after the visual grounding step has already succeeded.

\stopcontents[appendix]

\end{document}